%% file: main.tex
\documentclass{article}

\usepackage[preprint]{neurips_2026}

\usepackage[utf8]{inputenc}
\usepackage[T1]{fontenc}
\usepackage[colorlinks=true,linkcolor={blue!50!black},citecolor={blue!50!black},urlcolor={blue!50!black}]{hyperref}
\usepackage{url}
\usepackage{booktabs}
\usepackage{amsfonts}
\usepackage{nicefrac}
\usepackage{microtype}
\usepackage{xcolor}
\usepackage{amsmath,amssymb,amsthm}
\usepackage{graphicx}
\usepackage{bm}
\usepackage{tikz}

\newtheorem{theorem}{Theorem}
\newtheorem{corollary}{Corollary}
\newtheorem{proposition}{Proposition}
\newtheorem{lemma}{Lemma}[section]
\theoremstyle{remark}
\newtheorem{remark}{Remark}

\newcommand{\R}{\mathbb{R}}
\newcommand{\E}{\mathbb{E}}
\newcommand{\Var}{\mathrm{Var}}
\newcommand{\KL}{\mathrm{KL}}
\newcommand{\cS}{\mathcal{S}}
\newcommand{\cA}{\mathcal{A}}
\newcommand{\cZ}{\mathcal{Z}}
\newcommand{\cF}{\mathcal{F}}
\newcommand{\cG}{\mathcal{G}}
\newcommand{\pref}{\pi_{\mathrm{ref}}}
\newcommand{\Ph}{\widehat P}
\newcommand{\Vh}{\widehat V}
\newcommand{\Qh}{\widehat Q}
\newcommand{\pih}{\widehat\pi}
\newcommand{\SubOpt}{\mathrm{SubOpt}}
\DeclareMathOperator{\osc}{osc}

\input{fig_gadget.tex}
\hypersetup{pdftitle={Sharp Sample Complexity of KL-Regularized Reinforcement Learning},pdfauthor={Andrew Liu}}
\newcommand{\CodeAvailability}{The code is available from the author on request.}

\title{Sharp Sample Complexity of\\KL-Regularized Reinforcement Learning}

\author{Andrew Liu\\
Columbia University\\
\texttt{all2209@columbia.edu}}

\begin{document}

\maketitle

\begin{abstract}
KL-regularized reinforcement learning, the objective behind maximum-entropy RL and RL
from human feedback, is variational inference in the graphical model of ``control as
inference'': the regularized optimal policy is a posterior over trajectories, and the
suboptimality of a policy is $\beta$ times a KL divergence between path measures. However, it is
not known how many samples it takes to learn this posterior, even with a generative model.
Earlier $1/\epsilon$ guarantees scale with $H^5$ in the effective horizon $H$, and no lower bound for
MDPs depends on the temperature $\beta$. In this work, we show that for a discounted MDP with $K$
state-action pairs, the plug-in estimator finds a policy that is $\epsilon$-optimal from every initial
state with $\widetilde O(KH^3/(\epsilon\max\{\epsilon,\beta\})+KH)$ samples, and that no algorithm, even
one that chooses where to sample, can do with fewer when $\epsilon\lesssim\min\{H,H^2/\beta\}$. For
$\epsilon<\beta$ this is a $1/\epsilon$ rate with the same horizon dependence as unregularized RL. We
also find the exact asymptotic constant of the plug-in estimator, a Fisher-weighted variance of
the estimated soft $Q$-function. The estimator is locally asymptotically minimax among methods
that sample every pair equally often, and we give the allocation of samples that is optimal
among adaptive designs. Simulations on small MDPs match the rates and the constants.
\end{abstract}

\section{Introduction}

Many reinforcement learning (RL) methods optimize a regularized objective: the
expected discounted reward minus $\beta$ times the Kullback--Leibler divergence between
the policy and a reference policy $\pref$. Maximum-entropy RL and soft $Q$-learning
\citep{ziebart2010modeling,fox2016taming,haarnoja2017reinforcement,haarnoja2018soft}
take $\pref$ uniform; RL fine-tuning of language models takes $\pref$ to be the
pretrained model \citep{ouyang2022training}. The same objective arises from
probabilistic inference. In the ``control as inference'' graphical model
\citep{todorov2009efficient,rawlik2012stochastic,levine2018reinforcement}, each step
carries a binary optimality variable whose likelihood is $e^{r/\beta}$. With
deterministic dynamics, the posterior over trajectories given optimality is
$p(\tau)\propto p_{\mathrm{ref}}(\tau)e^{R(\tau)/\beta}$ and is generated by the soft-optimal
policy. With stochastic dynamics, the exact posterior changes the dynamics and becomes
risk-seeking, and the soft-optimal policy is the variational posterior that keeps the
true dynamics \citep{levine2018reinforcement}, and \citet{korbak2022rl} interpret RLHF in the
same way. In both cases the suboptimality of a policy $\pi$ equals $\beta$
times the KL divergence from the path law of $\pi$ to the path law of the soft-optimal
policy (Lemma~\ref{lem:pathkl}). So learning an $\epsilon$-optimal regularized policy is the same
problem as learning the posterior of control as inference to KL accuracy $\epsilon/\beta$.

For unregularized RL with a generative model the answer is known: $\Theta(KH^3/\epsilon^2)$
samples up to logarithmic factors \citep{azar2013minimax,agarwal2020model,li2024breaking}.
Regularization makes the objective strongly concave in the policy, which allows
$1/\epsilon$ rates. \citet{lan2022policy} and \citet{jia2026value} obtain
$\widetilde O(KH^5/(\beta\epsilon))$ with mirror-descent methods; \citet{jia2026value} also
give $\widetilde O(KH^3/\epsilon^2)$ for general convex regularizers, so the best previous guarantee
was $\min\{KH^3/\epsilon^2,KH^5/(\beta\epsilon)\}$. The plug-in estimator
inherits a $1/\epsilon$ rate from a sup-norm $Q$ bound, as \citet{tiapkin2023fast} remark, and working this out gives
$KH^4/(\beta\epsilon)$ (Section~\ref{sec:ideas}). In the online and offline settings the
horizon dependence is looser still \citep{tiapkin2023fast,tiapkin2024demonstration,zhao2025logarithmic,hao2026fast}.
Sharp two-regime rates are known for bandits ($H=1$) \citep{zhao2025sharp,ji2026optimal}.
However, for MDPs no lower bound that depends on $\beta$ is known, and \citet{tiapkin2023fast}
list the optimal rate as open.

In this work, we settle the generative-model case up to logarithmic factors for
$\epsilon\lesssim\min\{H,H^2/\beta\}$ (Table~\ref{tab:rates}). We show that with $N$ samples per pair
the plug-in estimator satisfies
$\max_s(V^*(s)-V^{\pih}(s))\lesssim\min\{\sqrt{H^3/N},\,H^3/(\beta N)\}$ up to logarithmic factors
once $N$ exceeds a burn-in of order $H$ (Theorem~\ref{thm:upper}), and that no algorithm does better
in this range, even one that chooses where to sample (Theorem~\ref{thm:lower}). In the fast regime
the horizon dependence is therefore $H^3$, as for unregularized RL, and for $\epsilon<\beta$ this
improves the best previous guarantee by a factor $\min\{H^2,\beta/\epsilon\}$. The main step is to
bound the across-action variance of the $Q$-error along the learner's own trajectories with a
martingale identity, instead of summing per-step worst cases (Section~\ref{sec:ideas}).

We also find the exact constant. For an initial distribution $\rho$, the risk of the plug-in
estimator is $\mathrm{tr}(W\Sigma_Q)/(2\beta N)+O(N^{-3/2})$, where $\Sigma_Q$ is the asymptotic
covariance of the estimated soft $Q$-function and $W$ weights it by the Fisher information of the
softmax policy along the optimal occupancy (Theorem~\ref{thm:asymp}). The estimator is locally
asymptotically minimax among methods that use $N$ samples per pair (Proposition~\ref{prop:uniform}).
When the samples may be allocated freely, sampling pair $z$ in proportion to $\sqrt{c_z}$ for
explicit weights $c_z$ is optimal, also among adaptive designs (Theorem~\ref{thm:design}). The
constant is at most of order $H^3/\beta$, and at most of order $H^2/\beta$ when no noisy pair can be
visited twice (Proposition~\ref{prop:revisit}).

\begin{table}[t]
\caption{Sample complexity (number of transitions) for an $\epsilon$-optimal policy for the
KL-regularized objective, up to logarithmic factors. $K$: state-action pairs. The episodic,
offline and bandit rows use their own definitions of $H$, $S$, $A$ ($C^\star$: single-policy
concentrability; $\eta=1/\beta$). A dash means that no lower bound depending on $\beta$ is known.
$^{\dagger}$For $\epsilon\lesssim\min\{H,H^2/\beta\}$.}
\label{tab:rates}
\centering
\footnotesize
\setlength{\tabcolsep}{5pt}
\begin{tabular}{llll}
\toprule
Reference & Setting & Upper bound & Lower bound\\
\midrule
\citet{azar2013minimax}; \citet{li2024breaking} & generative, $\beta=0$ & $KH^3/\epsilon^2$ & $KH^3/\epsilon^2$\\
\citet{lan2022policy} & generative & $KH^5/(\beta\epsilon)$ & --\\
\citet{jia2026value} & generative & $\min\{KH^3/\epsilon^2,\,KH^5/(\beta\epsilon)\}$ & --\\
\citet{tiapkin2023fast} (remark) & generative & $1/\epsilon$ rate, no explicit $H$ & --\\
\citet{tiapkin2024demonstration} & online, episodic & $H^5S^2A/(\beta\epsilon)$ & --\\
\citet{hao2026fast} & offline, episodic & $\eta H^7SAC^\star/\epsilon$ & --\\
\citet{zhao2025sharp,ji2026optimal} & bandits ($H=1$) & two regimes & matching\\
\midrule
\textbf{This paper} & generative & $KH^3/(\epsilon\max\{\epsilon,\beta\})$ & same$^{\dagger}$\\
\bottomrule
\end{tabular}
\end{table}

\section{Preliminaries}\label{sec:setting}

A discounted MDP has finite state space $\cS$, finite action sets $\cA_s$, the set
$\cZ=\{(s,a):a\in\cA_s\}$ of $K=|\cZ|$ state-action pairs, a transition kernel $P$,
known rewards $r:\cZ\to[0,1]$ and discount $\gamma\in[1/2,1)$; $H=1/(1-\gamma)$. A
reference policy $\pref$ gives positive probability to every action, and $\beta>0$ is the
temperature. The regularized value of a
policy $\pi$ is
\[
V^\pi(s)=\E^\pi_s\Big[\sum_{t\ge0}\gamma^t\Big(r(s_t,a_t)-\beta\KL\big(\pi(\cdot\mid s_t)\,\|\,\pref(\cdot\mid s_t)\big)\Big)\Big].
\]
The soft-optimal value and policy satisfy $Q^*=r+\gamma PV^*$,
$V^*(s)=\beta\log\sum_a\pref(a\mid s)e^{Q^*(s,a)/\beta}$ and
$\pi^*(a\mid s)=\pref(a\mid s)e^{(Q^*(s,a)-V^*(s))/\beta}$; then $0\le V^*\le H$ and $V^*\ge V^\pi$ for
every $\pi$. A \emph{generative model} returns independent draws $s'\sim P(\cdot\mid z)$ for
any queried pair $z$. The \emph{plug-in estimator} draws $N$ samples per pair (total
$T=KN$), forms the empirical kernel $\Ph$, and returns the soft-optimal policy $\pih$ of
the empirical MDP, i.e., of $(\Ph,r,\gamma,\beta,\pref)$. We write $\SubOpt(\pi)=\max_s(V^*(s)-V^\pi(s))$ for the worst-case
suboptimality over initial states and $\SubOpt_\rho(\pi)=V^*(\rho)-V^\pi(\rho)$ for an initial
distribution $\rho$. Since $\SubOpt_\rho\le\SubOpt$, we state the upper bound for $\SubOpt$ and the
lower bounds and exact constants for $\SubOpt_\rho$, whose expectation $\E\,\SubOpt_\rho(\pih)$ we call
the risk.
Finally $\iota=\log(4K(N+1)/\delta)$.

\begin{lemma}[suboptimality is a path KL; \citealp{nachum2017bridging,mei2020global}]\label{lem:pathkl}
For every policy $\pi$ and state $s$,
$V^*(s)-V^\pi(s)=\beta\,\E^\pi_s\sum_{t\ge0}\gamma^t\KL(\pi(\cdot\mid s_t)\|\pi^*(\cdot\mid s_t))$, which is
$\beta$ times the KL divergence between the laws of the trajectories of $\pi$ and $\pi^*$
from $s$, killed at rate $1-\gamma$.
\end{lemma}

\begin{remark}[exact versus variational inference]
Conditioning on optimality with stochastic dynamics gives the backup
$Q(s,a)=r(s,a)+\gamma\beta\log\E_{s'\sim P(\cdot\mid s,a)}e^{V(s')/\beta}$, an entropic-risk (risk-seeking)
recursion \citep{levine2018reinforcement}. Learning it requires $e^{\Omega(H/\beta)}$
samples, by the lower bounds for recursive entropic risk of
\citet{mortensen2025recursive} (see also \citealp{fei2020risk}). Appendix~\ref{app:misc}
works out a small example, the jackpot MDP, shown in Figure~\ref{fig:inference}(a). The
variational posterior studied here needs only polynomially many samples.
\end{remark}

\section{Main results}\label{sec:results}

\subsection{Minimax rates}

\begin{theorem}[upper bound]\label{thm:upper}
There are absolute constants $c_0,C$ such that for every $\delta\in(0,1)$ and $N\ge c_0H\iota$,
with probability at least $1-\delta$,
\[
\|\Qh^*-Q^*\|_\infty\le C\sqrt{\frac{H^3\iota}{N}}
\qquad\text{and}\qquad
\SubOpt(\pih)\le C\min\Big\{\sqrt{\frac{H^3\iota}{N}},\ \frac{H^3\iota}{\beta N}\Big\}.
\]
\end{theorem}

\begin{corollary}[sample complexity]\label{cor:samples}
For every $\epsilon\in(0,H]$, the plug-in estimator is $\epsilon$-optimal with probability at least
$1-\delta$ whenever $N\ge C^2H^3\iota/(\epsilon\max\{\epsilon,\beta\})+c_0H\iota$. Since $\iota$ grows only like
$\log N$, this holds for all large $N$, and $T=KN=\widetilde O(KH^3/(\epsilon\max\{\epsilon,\beta\})+KH)$
samples suffice.
\end{corollary}

\begin{theorem}[lower bounds]\label{thm:lower}
Let $\gamma\ge4/7$ and $\beta>0$. There is an absolute constant $c>0$ such that the following
hold.
(a) If $N\ge H/16$, there are two MDPs with five state-action pairs and a state $x$ such that
every estimator that uses $N$ samples per pair has, on one of them,
$\E[V^*(x)-V^{\pih}(x)]\ge c\min\{H^3/(\beta N),\sqrt{H^3/N}\}$.
(b) For $K\ge5$ and $\epsilon$ with $\epsilon\max\{\epsilon,\beta\}\le cH^2$, there is a family of MDPs
with at most $K$ pairs and an initial distribution $\rho$ such that every algorithm that
draws $T$ samples, possibly choosing each query after seeing the previous ones, and
guarantees $\E[V^*(\rho)-V^{\pih}(\rho)]\le\epsilon$ on the whole family must have
$T\ge cKH^3/(\epsilon\max\{\epsilon,\beta\})$.
\end{theorem}

Theorems~\ref{thm:upper} and~\ref{thm:lower} give the minimax sample complexity
$\widetilde\Theta(KH^3/(\epsilon\max\{\epsilon,\beta\}))$ for $\epsilon\lesssim\min\{H,H^2/\beta\}$. The two regimes meet at
$\epsilon\asymp\beta$, equivalently $N\asymp H^3/\beta^2$: once the $Q$-error falls below the temperature,
the loss becomes quadratic in it. The upper bound does not depend on $\pref$, which may put
arbitrarily small mass on some actions.

The regularized optimum is within $\beta H\log|\cA|$ of the unregularized one (for uniform
$\pref$), so using regularization only to approximate the unregularized optimum to accuracy
$\epsilon$ means $\beta\asymp\epsilon/(H\log|\cA|)$. That puts the problem in the slow regime and gives back
$KH^3/\epsilon^2$, consistent with \citet{kozuno2022kl}. At the other end, very strong
regularization needs no data:

\begin{proposition}\label{prop:trivial}
For every MDP, $\SubOpt(\pref)\le H^3/(8\beta)$.
\end{proposition}

So some restriction on $\epsilon$ is necessary.

\subsection{Exact constants and optimal designs}

Fix an instance and an initial distribution $\rho$. Let $P_{\pi^*}$ be the
$K\times K$ matrix $P_{\pi^*}[(s,a),(s',a')]=P(s'\mid s,a)\pi^*(a'\mid s')$, $M=(I-\gamma P_{\pi^*})^{-1}$,
$\sigma_z^2=\Var_{s'\sim P(\cdot\mid z)}V^*(s')$, and
$\Sigma_Q=\gamma^2M\,\mathrm{diag}(\sigma^2)M^\top$, the asymptotic covariance of
$\sqrt N(\Qh^*-Q^*)$. Let $d(s)=\sum_{t\ge0}\gamma^t\Pr^{\pi^*}_\rho(s_t=s)$ be the (unnormalized, total mass
$H$) discounted state occupancy of $\pi^*$,
$F_s=\mathrm{diag}(\pi^*(\cdot\mid s))-\pi^*(\cdot\mid s)\pi^*(\cdot\mid s)^\top$ the Fisher information of the
softmax family at $\pi^*(\cdot\mid s)$ in its natural parameter (up to $\beta^{-2}$), and
$W=\mathrm{blockdiag}_s(d(s)F_s)$.

\begin{theorem}[exact asymptotics]\label{thm:asymp}
As $N\to\infty$ with $N$ samples per pair:
(i) $N(V^*(\rho)-V^{\pih}(\rho))\to\frac{1}{2\beta}Z^\top WZ$ in distribution, $Z\sim\mathcal N(0,\Sigma_Q)$;
(ii) $N\,\E[V^*(\rho)-V^{\pih}(\rho)]\to\mathcal C:=\frac{1}{2\beta}\mathrm{tr}(W\Sigma_Q)$;
(iii) for every sequence of estimators that use $N$ samples per pair,
$\lim_{c\to\infty}\liminf_{N}N\sup_{\|\theta-\theta_0\|\le c/\sqrt N}\E_\theta[V^*_\theta(\rho)-V^{\widetilde\pi}_\theta(\rho)]\ge\mathcal C$,
where $\theta$ parametrizes kernels with the support of $P=P_{\theta_0}$.
\end{theorem}

The convergence in (ii) is uniform over a neighbourhood of the instance with error
$O(N^{-1/2})$ (Proposition~\ref{prop:uniform} in Appendix~\ref{app:uniform}), so the
plug-in estimator attains $\mathcal C$ uniformly over shrinking neighbourhoods; with (iii), it is
locally asymptotically minimax among estimators that use $N$ samples per pair.

\begin{theorem}[optimal allocation]\label{thm:design}
Let $c_z=\gamma^2\sigma_z^2[M^\top WM]_{zz}$, so that $\mathcal C=\frac1{2\beta}\sum_zc_z$.
(i) For every algorithm that draws $T$ samples adaptively,
$\lim_{c\to\infty}\liminf_{T}T\sup_{\|\theta-\theta_0\|\le c/\sqrt T}\E_\theta[V^*_\theta(\rho)-V^{\widetilde\pi}_\theta(\rho)]\ge\frac{1}{2\beta}\big(\sum_z\sqrt{c_z}\big)^2$.
(ii) The plug-in estimator with $w_zT$ samples at $z$ ($w_z>0$, $\sum_zw_z=1$) has
$T\,\E[V^*(\rho)-V^{\pih}(\rho)]\to\frac1{2\beta}\sum_zc_z/w_z$. If all $c_z>0$, the choice
$w_z\propto\sqrt{c_z}$ attains the bound in (i); in general, allocations with a small floor
come within any factor $1+\eta$ of it.
\end{theorem}

Unless every $c_z$ is zero (then both risks are $o(1/T)$, as for a deterministic kernel), uniform
sampling therefore loses a factor $K\sum_zc_z/(\sum_z\sqrt{c_z})^2\in[1,K]$. The factor is large when a
few pairs carry most of the influence, such as noisy transitions that the optimal policy
visits often.

\begin{proposition}[instance dependence]\label{prop:revisit}
Let $n_z=\sum_t\gamma^t\mathbb 1\{(s_t,a_t)=z\}$ be the discounted number of visits to $z$
under $\pi^*$. Then $\mathcal C\le\frac1\beta\max_{s_0}\sum_z\sigma_z^2\E^{\pi^*}_{s_0}[n_z^2]\le108H^3/\beta$,
and the bound improves to $108H^2/\beta$ if no pair with $\sigma_z>0$ can be visited twice in
a trajectory (e.g., in a time-indexed state space).
\end{proposition}

The $H^3$ of the minimax rate comes from repeated visits: an estimation error at a pair
that a trajectory revisits about $H$ times enters the value coherently each time. The hard
instances of Theorem~\ref{thm:lower} are built this way, from self-loops with exit
probability of order $1/H$ \citep{azar2013minimax}.

\section{Proof ideas}\label{sec:ideas}

All proofs are in the appendix; here we explain the upper bound, which is the main
technical step. Write $f=\Qh^*-Q^*$, $g=\Vh^*-V^*$,
$\kappa=\beta\KL(\pih\|\pi^*)$ and $\xi'=\gamma(\Ph-P)\Vh^*$. By Lemma~\ref{lem:pathkl},
$\SubOpt(\pih)=\max_s\E^{\pih}_s\sum_t\gamma^t\kappa(s_t)$, and since $\pih\propto\pi^*e^{f/\beta}$, a
Bennett-type bound gives $\kappa(s)\le\frac{3}{4\beta}\Var_{a\sim\pih(\cdot\mid s)}f(s,a)$ once $\|f\|\le\beta/2$
(Lemma~\ref{lem:klvar}).

\paragraph{Why the naive argument gives $H^4$.}
Bounding $\Var_{\pih}f\le\|f\|^2\lesssim H^3/N$ at every state and summing over the $H$
effective steps gives $\SubOpt\lesssim H\cdot H^3/(\beta N)$. This treats the errors at
consecutive states as unrelated---but $f$ solves a Bellman equation driven by the noise
$\xi'$, so they are not.

\paragraph{A martingale identity.}
Two exact identities hold (Lemma~\ref{lem:identities}): $f=\xi'+\gamma Pg$ and
$g=\pih f-\kappa=\pi^*f+\kappa'$ with $\kappa'=\beta\KL(\pi^*\|\pih)\ge0$. So $g$ is the value of
$\pih$ for the reward $\xi'$ and the state cost $\kappa$, and $f$ is its $Q$-function.
Along a trajectory of $\pih$, write $D_t=f(s_t,a_t)-\pih f(s_t)$ (action noise) and
$E_{t+1}=g(s_{t+1})-Pg(s_t,a_t)$ (transition noise). Then
\[
\sum_t\gamma^t\big(\xi'(s_t,a_t)-\kappa(s_t)\big)-g(s_0)=\sum_t\gamma^t\big(D_t+\gamma E_{t+1}\big),
\]
and orthogonality of martingale increments gives
$\sum_t\gamma^{2t}\E[\Var_{\pih}f(s_t)]\le3\big(\E(\sum_t\gamma^t\xi'_t)^2+\E(\sum_t\gamma^t\kappa(s_t))^2+g(s_0)^2\big)$.
The left side is exactly what the suboptimality needs (after a comparison between
$\gamma^t$ and $\gamma^{2t}$ weights, Lemma~\ref{lem:resolvent}). On the right, the second term is
at most $2\SubOpt^2$ by a potential inequality (Lemma~\ref{lem:gn}), because the
expected remaining sum of $\kappa$ from any time is the suboptimality at the current
state, and $\E(\sum_t\gamma^t\xi'_t)^2\le H\,\E\sum_t\gamma^t\xi'^2_t$.

\paragraph{A closed system.}
With absorbing-MDP decoupling \citep{agarwal2020model} adapted to soft Bellman
operators, Bernstein's inequality gives
$\xi'^2\lesssim\frac{\iota}{N}(\Var_PV^*+\Var_Pg)$ at every pair (Lemma~\ref{lem:noise}). A soft
total variance lemma bounds $\E\sum_t\gamma^t\Var_PV^*(s_t,a_t)$ by $O((H+\SubOpt)^2)$ along any
policy (Lemma~\ref{lem:tv}); the extra difficulty compared with the unregularized case
is the random KL penalty, whose sum is controlled by the same potential inequality.
The term $\Var_Pg$ is the transition noise of $g$ in the identity above, so it is
bounded by the same right side. This yields inequalities between the noise second
moments along $\pih$ and $\pi^*$, $\|g\|$ and $\SubOpt$ that close once $N\gtrsim H\iota$
(Lemma~\ref{lem:closed}) and give $\E\sum_t\gamma^t\xi'^2_t\lesssim H^2\iota/N$, $\|g\|,\|f\|,\SubOpt\lesssim\sqrt{H^3\iota/N}$, and finally
$\SubOpt\lesssim\frac1\beta(H^3\iota/N+\SubOpt^2)$. With the slow rate for $\SubOpt$, the last term
can be absorbed once $N\gtrsim H^3\iota/\beta^2$.

\paragraph{Lower bound.}
A decision state with two actions leads to two self-loops with stay probabilities
$p_0\pm\Delta$, $1-p_0\asymp1/H$ (Figure~\ref{fig:gadget}a). The $Q$-gap is about $H^2\Delta$, and for any learned action
probability the regularized losses under the two hypotheses sum to at least
$2\beta\log\cosh(\Delta_Q/(2\beta))$ (Lemma~\ref{lem:twopoint}). Since $\log\cosh v\asymp\min\{v^2,v\}$,
one argument covers both regimes and the transition between them. With $\Delta\asymp(HN)^{-1/2}$ the hypotheses are indistinguishable and the
loss is $\min\{H^3/(\beta N),\sqrt{H^3/N}\}$; Assouad's lemma with the chain rule for
adaptive sampling extends this to the total budget.

\begin{figure}[t]
\centering
\GadgetLoop\hspace{1.4cm}\GadgetChain
\caption{The two-arm gadget of Theorem~\ref{thm:lower}(a). (a) From the decision state $x$, action $i$
leads to a state $y_i$ with reward $1$, which the process leaves for the absorbing state $w$ (reward $0$)
with probability $1-p_i$; here $p_{1,2}=p_0\pm\Delta$ and $1-p_0\asymp1/H$. (b) The time-indexed version
used in Section~\ref{sec:experiments}: each loop is unrolled into a chain of distinct states with reward
$1$, each moving on with probability $p_i$ and otherwise to $w$, so no state is visited twice.}
\label{fig:gadget}
\end{figure}

\section{Simulations}\label{sec:experiments}

\begin{figure}[t]
\centering
\includegraphics[width=\linewidth]{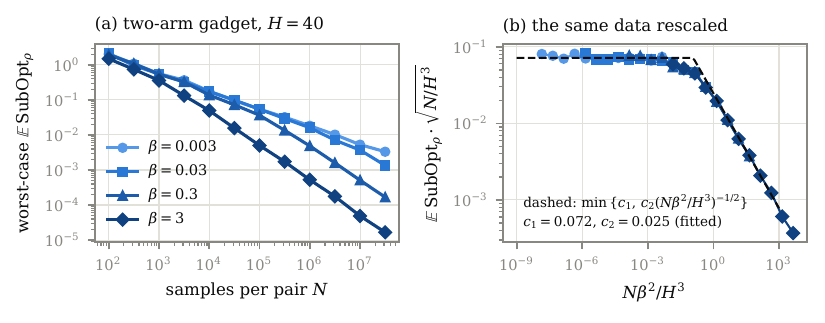}
\caption{Rates in $N$ (Theorems~\ref{thm:upper} and~\ref{thm:lower}). (a) Expected suboptimality of the
plug-in estimator on the two-arm gadget with $H=40$, worst case over 13 values of the gap $\Delta$.
Each point averages 900 datasets; standard errors are smaller than the markers. (b) The same data,
$\E\,\SubOpt_\rho\cdot\sqrt{N/H^3}$ against $N\beta^2/H^3$. The dashed line is
$\min\{c_1,c_2(N\beta^2/H^3)^{-1/2}\}$, with $c_1$ and $c_2$ read off the two ends of the range; near the
bend the data lie below it.}
\label{fig:rates}
\end{figure}

\begin{figure}[t]
\centering
\includegraphics[width=\linewidth]{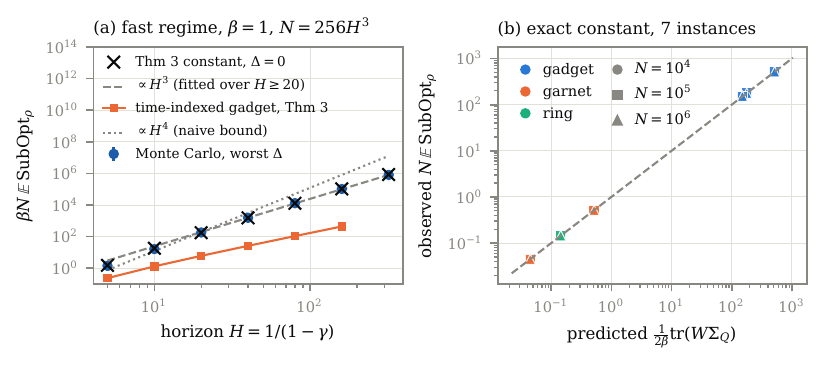}
\caption{(a) Horizon dependence in the fast regime ($\beta=1$, $N=256H^3$). Dots: $\beta N\,\E\,\SubOpt_\rho$
at the worst gap $\Delta$, estimated from 900 datasets. Crosses: the constant of Theorem~\ref{thm:asymp}
at $\Delta=0$. Squares: the same constant for the time-indexed version of the gadget. (b) Measured
$N\,\E\,\SubOpt_\rho$ against the constant of Theorem~\ref{thm:asymp} on seven instances at three sample
sizes (3000 datasets each). Standard errors are smaller than the markers.}
\label{fig:horizon}
\end{figure}

We ran small simulations to see how the rates and constants show up at moderate sample sizes.
Throughout we report $\SubOpt_\rho$. On the gadget of Theorem~\ref{thm:lower}(a), drawn in
Figure~\ref{fig:gadget}, $\rho$ is the decision state, so $\SubOpt_\rho=\SubOpt$. Appendix~\ref{app:experiments}
gives the details.

\subsection{Both regimes appear at moderate sample sizes}
We first checked the rate in $N$ on the gadget with $H=40$ and four temperatures. The worst-case risk
over the gap $\Delta$ falls like $N^{-1/2}$ while $N\lesssim H^3/\beta^2$ and like $N^{-1}$ after that
(Figure~\ref{fig:rates}a), and after rescaling the four curves fall on one (Figure~\ref{fig:rates}b). Across
$10^2\le N\le3\cdot10^7$ the risk stays within a factor $3.6$ of $\min\{\sqrt{H^3/N},H^3/(\beta N)\}$, with
constants in front far smaller than those of Theorem~\ref{thm:upper}.

\subsection{The constant grows like $H^3$, and like $H^2$ without revisits}
The fast regime is where our bound improves on earlier ones, so we followed it in $H$, with $\beta=1$
and $N=256H^3$. As expected, $\beta N\,\E\,\SubOpt_\rho$ grows like $H^{3.02}$ for $20\le H\le320$,
not like the $H^4$ of the naive argument (Figure~\ref{fig:horizon}a). At $\Delta=0$ the risk has a closed
form, and it agrees with the constant of Theorem~\ref{thm:asymp} to within $0.07\%$ at every $H$
(Table~\ref{tab:horizon}). In the slow regime ($\beta=10^{-4}$, $N=64H$) the ratio $\E\,\SubOpt_\rho/\sqrt{H^3/N}$
grows only like $H^{0.11}$.

Proposition~\ref{prop:revisit} ties the $H^3$ to repeated visits. To test this, we unrolled each loop of
the gadget into a chain of distinct states with the same transition probabilities
(Figure~\ref{fig:gadget}b). The exact constant then grows like $H^{2.06}$, against $H^{3.07}$
for the gadget with loops over the same range of $H$ (Figure~\ref{fig:horizon}a).

\subsection{The asymptotic constant matches the measured risk}
Theorem~\ref{thm:asymp} gives the constant itself, so we compared $\frac1{2\beta}\mathrm{tr}(W\Sigma_Q)$ with the
measured $N\,\E\,\SubOpt_\rho$ at $N=10^6$ on seven instances whose constants span four orders of
magnitude: gadgets, random ``Garnet'' MDPs with random reference policies, and a 100-state ring. The two
agree to within $2.1\%$ (Figure~\ref{fig:horizon}b and Table~\ref{tab:constants}). On the gadget the
constant carries the Fisher weight $\pi^*(1\mid x)\,\pi^*(2\mid x)$ of the decision state $x$, so it decays
exponentially in the log-odds gap at $x$; the measured risk follows it (Figure~\ref{fig:inference}b).

\subsection{Sampling in proportion to $\sqrt{c_z}$ lowers the risk}
On three instances, the allocation of Theorem~\ref{thm:design} lowers the risk by up to a factor
$2.57$ relative to uniform sampling, close to the gain the theorem gives, and the risk of each
design is close to its constant (Table~\ref{tab:allocation}).

\begin{table}[t]
\caption{Uniform sampling versus the allocation $w_z\propto\sqrt{c_z}$ of Theorem~\ref{thm:design}, with
budget $T=K\cdot10^5$: the constant $\frac1{2\beta}\sum_zc_z/w_z$ and the measured $T\,\E\,\SubOpt_\rho$ (3000
datasets per entry, $\pm$ one standard error). The gadget has $\beta=1$, the Garnets $\gamma=0.95$ and
$\beta=0.5$. The gain is the ratio of the uniform to the optimal risk.}
\label{tab:allocation}
\centering\small
\begin{tabular}{lrrrrr}
\toprule
instance & $K$ & design & constant & measured & gain (theory / measured) \\
\midrule
gadget, $H=20$ & 5 & uniform & 888.4 & 899.8 $\pm$ 23 & 2.50 / 2.57 \\
gadget, $H=20$ & 5 & optimal & 355.4 & 350.6 $\pm$ 9.1 &  \\
Garnet, $S=10$, $A=4$ & 40 & uniform & 12.9 & 12.95 $\pm$ 0.1 & 1.50 / 1.52 \\
Garnet, $S=10$, $A=4$ & 40 & optimal & 8.625 & 8.539 $\pm$ 0.052 &  \\
Garnet, $S=20$, $A=2$ & 40 & uniform & 13.62 & 13.92 $\pm$ 0.21 & 2.19 / 2.23 \\
Garnet, $S=20$, $A=2$ & 40 & optimal & 6.23 & 6.247 $\pm$ 0.051 &  \\
\bottomrule
\end{tabular}
\end{table}

\subsection{Exact inference needs exponentially many samples}
Figure~\ref{fig:inference}(a) shows the jackpot MDP for $H/\beta\le20$. Proposition~\ref{prop:ei}
needs $H/\beta\ge32$, but the separation is already clear at these values. The plug-in estimate of the
exact-inference posterior stays far from the truth until $N$ is of order $e^{H/(2\beta)}$, while the error
of the variational policy is below $0.1$ once $N\ge8$.

\section{Related work}\label{sec:related}

\paragraph{Generative-model RL.} Minimax rates for unregularized discounted MDPs are
due to \citet{azar2013minimax} and were extended to the full range of sample sizes by
\citet{agarwal2020model} and \citet{li2024breaking}. Model-free methods reach them with variance
reduction \citep{sidford2018near,wainwright2019variance}, while plain $Q$-learning needs $H^4$
\citep{li2024qlearning}. We extend the decoupling argument of \citet{agarwal2020model} to soft
Bellman operators and combine the hard instance of \citet{azar2013minimax} with a two-point
argument for the regularized loss.

\paragraph{Regularized MDPs.} Regularization speeds up planning
\citep{geist2019theory,cen2022fast} and is used as an algorithmic device for the
unregularized objective \citep{kozuno2022kl}. For learning the regularized objective itself,
mirror-descent methods give $1/\epsilon$ rates with a generative model
\citep{lan2022policy,jia2026value}, and later work covers the online and offline settings
\citep{tiapkin2023fast,tiapkin2024demonstration,zhao2025logarithmic,hao2026fast}; Table~\ref{tab:rates}
compares the rates. The closest results are for bandits, where \citet{zhao2025sharp} and
\citet{ji2026optimal} prove sharp two-regime rates with matching lower bounds. Our results give the
analogue for MDPs, where the horizon enters. \citet{hu2021fast} show fast regret rates for plug-in
policies under margin conditions.

\paragraph{Control as inference.} The inference view goes back to
\citet{todorov2009efficient}, \citet{rawlik2012stochastic} and
\citet{ziebart2010modeling}; see \citet{levine2018reinforcement} for a review and
\citet{neu2017unified} for its relation to entropy-regularized MDPs. Its consequences
for exploration were studied by \citet{odonoghue2020making} and
\citet{tarbouriech2023probabilistic}. Exact inference leads to the entropic-risk recursion
studied by \citet{fei2020risk} and \citet{mortensen2025recursive}. We use the inference view to pose
the question, and its variational identities in the proofs (the Gibbs variational principle
and the two forms of $g$ in Section~\ref{sec:ideas}).

\paragraph{Instance-dependent and asymptotic analysis.} \citet{li2023statistical} prove a
central limit theorem for entropy-regularized $Q$-learning with the covariance
$\Sigma_Q$; \citet{khamaru2020td,khamaru2021instance} give instance-optimal bounds for value
estimation. Theorem~\ref{thm:asymp} concerns the suboptimality of the plug-in policy, which is
second order in the estimation error. Its proof combines the second-order delta method
\citep{vandervaart1998asymptotic} with the van Trees inequality \citep{gill1995applications};
the allocation of Theorem~\ref{thm:design} has the form of Neyman's allocation in stratified
sampling \citep{neyman1934two}. Soft-optimal (logit) policies also appear in
dynamic discrete choice models in econometrics \citep{rust1987optimal,mai2020relation}, where
the goal is to estimate rewards rather than to learn a policy.

\section{Discussion}\label{sec:discussion}

Two readings of the results seem worth stating. The first is about control as inference. The exact
posterior---which lets the optimality variables bend the dynamics---needs a number of samples
exponential in $H/\beta$, while the variational posterior, which keeps the true dynamics, is learned at
the rates above. The usual preference for the variational form, which \citet{levine2018reinforcement}
motivates by the risk-seeking behaviour of exact inference, thus has a statistical side as well. Second,
Theorem~\ref{thm:upper} is a worst-case bound, and on typical instances the exact constant is far below
it: on the Garnet instances of Section~\ref{sec:experiments} it is several orders of magnitude smaller
than $H^3/\beta$. We suspect that a finite-sample bound in terms of the revisit moments of
Proposition~\ref{prop:revisit} holds, which would show this gap outside the asymptotic regime.

\subsection{Limitations}\label{sec:limitations}
Our results are for tabular MDPs with a generative model and known rewards. This setting isolates the
cost of learning the dynamics, but it leaves out exploration and function approximation, where the
data depend on the learner's own policy; the online setting is the natural next case. When
$\beta\gtrsim H$ the lower bound also leaves the window $H^2/\beta\lesssim\epsilon\lesssim H^3/\beta$ open. There
the hard instances cannot produce a loss above order $H^2/\beta$ and the upper bound is set by the
burn-in, so closing the window needs a different hard instance or a smaller burn-in. The optimal
allocation of Theorem~\ref{thm:design} needs the weights $c_z$, which depend on the unknown instance and
would have to be estimated. Finally, the constants of Theorem~\ref{thm:upper} are far from tight (its
logarithmic factor also includes a $\log N$ from the decoupling argument), so most of our simulated
sample sizes lie below the burn-in it requires.

\subsection{Conclusion}\label{sec:conclusion}

Learning the posterior of control as inference to KL accuracy $\epsilon/\beta$ takes
$\widetilde\Theta(KH^3/(\epsilon\max\{\epsilon,\beta\}))$ samples from a generative model when $\epsilon\lesssim\min\{H,H^2/\beta\}$,
and the plug-in estimator attains it with a constant that is locally optimal among methods that sample
every pair equally often. Regularization therefore buys a $1/\epsilon$ rate at no cost in the horizon.
The right horizon dependence is $H^3$ for the regularized problem, as for the unregularized one, and the
$H^4$ and $H^5$ of earlier bounds come from the analyses. The exact constant also shows that samples
matter most at the noisy transitions that the optimal policy keeps coming back to, in proportion to
$\sqrt{c_z}$. Natural next steps are a two-stage allocation that estimates
these weights from a pilot sample, the online setting, where \citet{tiapkin2023fast} conjecture the analogous
rate for best-policy identification, and offline data, where the best known horizon dependence is
$H^7$ \citep{hao2026fast}.

\bibliographystyle{plainnat}
\bibliography{refs}

\appendix
\input{app_prelim.tex}
\input{app_upper.tex}
\input{app_lower.tex}
\input{app_asymp.tex}
\input{app_misc.tex}

\end{document}

%% file: fig_gadget.tex
\tikzset{
  gstate/.style={circle, draw, minimum size=7mm, inner sep=0pt, font=\small},
  gsmall/.style={circle, draw, minimum size=5mm, inner sep=0pt},
  glabel/.style={font=\scriptsize},
}
\newcommand{\GadgetLoop}{%
\begin{tikzpicture}[>=stealth, semithick, baseline=(x.base)]
  \node[gstate] (x) at (0,0) {$x$};
  \node[gstate] (y1) at (1.9,0.8) {$y_1$};
  \node[gstate] (y2) at (1.9,-0.8) {$y_2$};
  \node[gstate] (w) at (3.8,0) {$w$};
  \draw[->] (x) -- node[glabel, above left=-2pt] {$a=1$} (y1);
  \draw[->] (x) -- node[glabel, below left=-2pt] {$a=2$} (y2);
  \draw[->] (y1) edge[loop above, min distance=7mm] node[glabel, above] {$p_1$} (y1);
  \draw[->] (y2) edge[loop below, min distance=7mm] node[glabel, below] {$p_2$} (y2);
  \draw[->] (y1) -- node[glabel, above right=-2pt] {$1-p_1$} (w);
  \draw[->] (y2) -- node[glabel, below right=-2pt] {$1-p_2$} (w);
  \node[glabel] at (-0.2,1.5) {(a)};
\end{tikzpicture}}
\newcommand{\GadgetChain}{%
\begin{tikzpicture}[>=stealth, semithick, baseline=(x.base)]
  \node[gstate] (x) at (0,0) {$x$};
  \foreach \row/\y/\p/\pos in {1/0.8/p_1/above, 2/-0.8/p_2/below} {
    \node[gsmall] (c\row1) at (1.4,\y) {};
    \node[gsmall] (c\row2) at (2.4,\y) {};
    \node[gsmall] (c\row3) at (3.4,\y) {};
    \node[glabel] (c\row4) at (4.2,\y) {$\cdots$};
    \draw[->] (x) -- (c\row1);
    \draw[->] (c\row1) -- node[glabel, \pos] {$\p$} (c\row2);
    \draw[->] (c\row2) -- node[glabel, \pos] {$\p$} (c\row3);
    \draw[->] (c\row3) -- (c\row4);
  }
  \node[glabel] at (-0.2,1.5) {(b)};
\end{tikzpicture}}

%% file: app_prelim.tex
\section{Preliminaries}\label{app:prelim}

\paragraph{Notation.}
$\cZ=\{(s,a):s\in\cS,a\in\cA_s\}$ is the set of state-action pairs, $K=|\cZ|$, and
$H=1/(1-\gamma)$ with $\gamma\in[1/2,1)$. For $v:\cS\to\R$ and $z\in\cZ$ we write
$Pv(z)=\sum_{s'}P(s'\mid z)v(s')$ and $\Var_P v(z)=P(v^2)(z)-(Pv(z))^2$; for
$q:\cZ\to\R$ and a policy $\pi$, $(\pi q)(s)=\sum_{a}\pi(a\mid s)q(s,a)$ and
$\Var_\pi q(s)=\pi(q^2)(s)-(\pi q(s))^2$. We write $\KL_\pi(s)=\KL(\pi(\cdot\mid s)\,\|\,\pref(\cdot\mid s))$.
For a policy $\pi$ and a kernel $P'$, $\E^{\pi,P'}_s$ denotes expectation over the trajectory
$(s_t,a_t)_{t\ge0}$ with $s_0=s$, $a_t\sim\pi(\cdot\mid s_t)$, $s_{t+1}\sim P'(\cdot\mid s_t,a_t)$;
we drop $P'$ when $P'=P$, the true kernel. $\cF_t=\sigma(s_0,a_0,\dots,s_t)$ and
$\cG_t=\cF_t\vee\sigma(a_t)$. Norms $\|\cdot\|$ are sup norms. The regularized Bellman
operator of $\pi$ under $P'$ is
\[
T^\pi_{P'}v(s)=\pi\big(r+\gamma P'v\big)(s)-\beta\KL_\pi(s),
\]
its fixed point is $V^\pi_{P'}$, and $Q^\pi_{P'}=r+\gamma P'V^\pi_{P'}$. The soft Bellman
operator is $T_{P'}v(s)=\beta\log\sum_a\pref(a\mid s)\exp\big((r(s,a)+\gamma P'v(s,a))/\beta\big)$.
For $q:\cZ\to\R$ we write $\pi_q(a\mid s)\propto\pref(a\mid s)e^{q(s,a)/\beta}$.

\begin{lemma}[Gibbs variational principle]\label{lem:gibbs}
For a state $s$, $q:\cA_s\to\R$ and any distribution $\mu$ on $\cA_s$,
\[
\beta\log\sum_a\pref(a\mid s)e^{q(a)/\beta}=\sum_a\mu(a)q(a)-\beta\KL(\mu\|\pref(\cdot\mid s))+\beta\KL(\mu\|\pi_q(\cdot\mid s)).
\]
\end{lemma}
\begin{proof}
Write $Z=\sum_a\pref(a\mid s)e^{q(a)/\beta}$, so $\log\pi_q(a\mid s)=\log\pref(a\mid s)+q(a)/\beta-\log Z$.
Then $\KL(\mu\|\pi_q)=\KL(\mu\|\pref)-\mu q/\beta+\log Z$; multiply by $\beta$ and rearrange.
\end{proof}

\begin{lemma}[basic properties]\label{lem:basic}
(i) $T_{P'}$ and $T^\pi_{P'}$ are monotone, satisfy $T(v+c\mathbf 1)=Tv+\gamma c\mathbf 1$, and are
$\gamma$-contractions in sup norm; their fixed points $V^*_{P'}$ and $V^\pi_{P'}$ exist and are unique.
(ii) $\pi^*_{P'}:=\pi_{Q^*_{P'}}$ with $Q^*_{P'}=r+\gamma P'V^*_{P'}$ satisfies $V^{\pi^*_{P'}}_{P'}=V^*_{P'}$.
(iii) If $r\in[0,1]$ then $0\le V^{\pref}_{P'}\le V^*_{P'}\le H$ and $0\le Q^*_{P'}\le H$.
\end{lemma}
\begin{proof}
(i) Monotonicity and the translation property are immediate (log-sum-exp is monotone and
commutes with constants); together they give the contraction. (ii) By
Lemma~\ref{lem:gibbs} with $\mu=\pi^*_{P'}(\cdot\mid s)$, $T_{P'}V^*=T^{\pi^*}_{P'}V^*$, so $V^*$
is the fixed point of $T^{\pi^*}_{P'}$. (iii) $\KL_{\pref}=0$ and $r\ge0$ give $V^{\pref}\ge0$;
Lemma~\ref{lem:gibbs} gives $T_{P'}v\ge T^{\pi}_{P'}v$ for every $\pi$, hence $V^*\ge V^\pi$ for
all $\pi$; and $T_{P'}v\le 1+\gamma\|v\|$ (take $\mu=\pi_q$ and use $\KL\ge0$) gives $V^*\le H$.
Finally $Q^*=r+\gamma P'V^*\in[0,1+\gamma H]=[0,H]$.
\end{proof}

\begin{lemma}[Lemma~\ref{lem:pathkl} restated]\label{lem:pdl}
For every policy $\pi$, kernel $P'$ and state $s$,
\[
V^*_{P'}(s)-V^\pi_{P'}(s)=\E^{\pi,P'}_s\sum_{t\ge0}\gamma^t\,\beta\KL\big(\pi(\cdot\mid s_t)\,\|\,\pi^*_{P'}(\cdot\mid s_t)\big).
\]
Equivalently, it is $\beta$ times the KL divergence between the laws of the trajectories of $\pi$
and $\pi^*_{P'}$ started at $s$ and killed at each step with probability $1-\gamma$.
\end{lemma}
\begin{proof}
By Lemma~\ref{lem:gibbs} with $\mu=\pi(\cdot\mid s)$ and $q=Q^*(s,\cdot)$,
$V^*(s)=\pi Q^*(s)-\beta\KL_\pi(s)+\beta\KL(\pi\|\pi^*)(s)$. Hence
$\E^{\pi}[V^*(s_t)-\gamma V^*(s_{t+1})-r(s_t,a_t)+\beta\KL_\pi(s_t)\mid\cF_t]=\beta\KL(\pi\|\pi^*)(s_t)$.
Multiply by $\gamma^t$, sum over $t$ (the sums converge absolutely because all terms are
bounded) and use $V^\pi(s)=\E^\pi_s\sum_t\gamma^t(r(s_t,a_t)-\beta\KL_\pi(s_t))$. For the path
form, the log-likelihood ratio of a killed trajectory $(s_0,a_0,\dots,s_\tau,a_\tau)$ is
$\sum_{t\le\tau}\log(\pi/\pi^*)(a_t\mid s_t)$ and $\Pr(\tau\ge t)=\gamma^t$.
\end{proof}

\begin{lemma}[KL between Gibbs policies]\label{lem:klvar}
Let $p,q$ be distributions on a finite set with $p\propto q\,e^{f/\beta}$, and suppose
$\osc(f)=\max f-\min f\le b\beta$. Then $\beta\KL(p\|q)\le\phi(b)\Var_p(f)/\beta$, where
$\phi(b)=(e^b-1-b)/b^2$. In particular $\phi(1)=e-2<3/4$.
\end{lemma}
\begin{proof}
$q=p\,e^{-f/\beta}/\E_p e^{-f/\beta}$, so $\KL(p\|q)=\E_p[f]/\beta+\log\E_p e^{-f/\beta}=\log\E_p e^X$
with $X=-(f-\E_pf)/\beta$, $\E_pX=0$, $X\le b$. Since $x\mapsto(e^x-1-x)/x^2$ is increasing,
$e^x\le1+x+\phi(b)x^2$ for $x\le b$, so $\log\E_pe^X\le\E_p[e^X-1]\le\phi(b)\E_pX^2=\phi(b)\Var_p(f)/\beta^2$.
\end{proof}

\begin{lemma}[a potential inequality]\label{lem:gn}
Let $x_t\ge0$ be $\cF_t$-measurable and suppose $\E[\sum_{t\ge k}\gamma^{t-k}x_t\mid\cF_k]\le c$
almost surely for every $k\ge0$. Then $\E[(\sum_{t\ge0}\gamma^tx_t)^2]\le 2c\,\E[\sum_{t\ge0}\gamma^tx_t]\le 2c^2$.
\end{lemma}
\begin{proof}
$(\sum_t\gamma^tx_t)^2\le2\sum_t\gamma^tx_t\sum_{t'\ge t}\gamma^{t'}x_{t'}$ (each off-diagonal pair once
with factor 2, the diagonal with factor 2). Conditioning on $\cF_t$,
$\E[\gamma^tx_t\sum_{t'\ge t}\gamma^{t'}x_{t'}]=\E[\gamma^{2t}x_t\,\E[\sum_{t'\ge t}\gamma^{t'-t}x_{t'}\mid\cF_t]]\le c\,\E[\gamma^tx_t]$.
Sum over $t$.
\end{proof}

\begin{lemma}[comparison of resolvents]\label{lem:resolvent}
Let $\mathcal P$ be a stochastic matrix on a finite set and $x\ge0$. Then
$\|(I-\gamma\mathcal P)^{-1}x\|\le(1+\gamma)\|(I-\gamma^2\mathcal P)^{-1}x\|$. In trajectory form:
$\max_{s}\E_s\sum_t\gamma^tx(s_t)\le(1+\gamma)\max_s\E_s\sum_t\gamma^{2t}x(s_t)$ for a Markov chain $(s_t)$.
\end{lemma}
\begin{proof}
Let $u=(I-\gamma\mathcal P)^{-1}x\ge0$ and $v=(I-\gamma^2\mathcal P)^{-1}x$. Then
$(I-\gamma^2\mathcal P)u=x+\gamma(1-\gamma)\mathcal Pu$, so $u=v+\gamma(1-\gamma)(I-\gamma^2\mathcal P)^{-1}\mathcal Pu$.
The matrix $(I-\gamma^2\mathcal P)^{-1}\mathcal P$ is nonnegative with row sums $1/(1-\gamma^2)$, so
$\|u\|\le\|v\|+\frac{\gamma}{1+\gamma}\|u\|$.
\end{proof}

\begin{lemma}[martingale variance identity]\label{lem:mvi}
Let $P'$ be a kernel, $\pi$ a policy, $\rho:\cZ\to\R$, $c:\cS\to\R$ and $v:\cS\to\R$ bounded with
$v=\pi(\rho+\gamma P'v)-c$. Put $q=\rho+\gamma P'v$, $A_t=q(s_t,a_t)-\pi q(s_t)$ and
$E_{t+1}=v(s_{t+1})-P'v(s_t,a_t)$. Along trajectories of $(\pi,P')$ from $s_0$,
\[
\sum_t\gamma^t\big(\rho(s_t,a_t)-c(s_t)\big)-v(s_0)=\sum_t\gamma^t\big(A_t+\gamma E_{t+1}\big),
\]
and
$\E\big[\big(\sum_t\gamma^t(\rho(s_t,a_t)-c(s_t))-v(s_0)\big)^2\big]=\sum_t\gamma^{2t}\E\big[\Var_\pi q(s_t)+\gamma^2\Var_{P'}v(s_t,a_t)\big]$.
In particular $v(s_0)=\E\sum_t\gamma^t(\rho(s_t,a_t)-c(s_t))$.
\end{lemma}
\begin{proof}
$v(s_t)=\pi q(s_t)-c(s_t)=\rho(s_t,a_t)-c(s_t)+\gamma v(s_{t+1})-\gamma E_{t+1}-A_t$; multiply by
$\gamma^t$ and sum (telescoping, using boundedness). The sequence
$A_0,\gamma E_1,\gamma A_1,\gamma^2E_2,\dots$ is a bounded martingale difference sequence for the
interleaved filtration $\cF_0\subset\cG_0\subset\cF_1\subset\cdots$ ($\E[A_t\mid\cF_t]=0$,
$\E[E_{t+1}\mid\cG_t]=0$), so its increments are uncorrelated and
$\E[A_t^2\mid\cF_t]=\Var_\pi q(s_t)$, $\E[E_{t+1}^2\mid\cG_t]=\Var_{P'}v(s_t,a_t)$.
\end{proof}

\begin{lemma}[soft total variance]\label{lem:tv}
Let $\pi$ be any policy and $\Delta_\pi=\max_s(V^*(s)-V^\pi(s))$ (true kernel). For every $s_0$,
\begin{gather*}
\sum_t\gamma^{2t}\E^{\pi}_{s_0}\big[\Var_\pi Q^*(s_t)+\gamma^2\Var_PV^*(s_t,a_t)\big]\le18(H+\Delta_\pi)^2,\\
\max_{s}\E^\pi_{s}\sum_t\gamma^t\Var_PV^*(s_t,a_t)\le108(H+\Delta_\pi)^2.
\end{gather*}
\end{lemma}
\begin{proof}
Let $\kappa_\pi(s)=\beta\KL(\pi\|\pi^*)(s)\ge0$. By Lemma~\ref{lem:gibbs},
$V^*=\pi(r+\gamma PV^*)-(\beta\KL_\pi-\kappa_\pi)$, so Lemma~\ref{lem:mvi} applies with $\rho=r$,
$c=\beta\KL_\pi-\kappa_\pi$, $v=V^*$, $q=Q^*$. The left side of the first claim equals
$\E[(R-L+K'-V^*(s_0))^2]$ with $R=\sum_t\gamma^tr(s_t,a_t)\in[0,H]$,
$L=\sum_t\gamma^t\beta\KL_\pi(s_t)\ge0$ and $K'=\sum_t\gamma^t\kappa_\pi(s_t)\ge0$. Writing the
argument as $(R-H/2)-L+K'+(H/2-V^*(s_0))$ and using $V^*\in[0,H]$,
$\E[(\cdot)^2]\le4(H^2/4+\E L^2+\E K'^2+H^2/4)$. By Lemma~\ref{lem:pdl} and the Markov property,
$\E[\sum_{t\ge k}\gamma^{t-k}\kappa_\pi(s_t)\mid\cF_k]=V^*(s_k)-V^\pi(s_k)\le\Delta_\pi$, so
Lemma~\ref{lem:gn} gives $\E K'^2\le2\Delta_\pi^2$. Also
$\E[\sum_{t\ge k}\gamma^{t-k}\beta\KL_\pi(s_t)\mid\cF_k]=\E[\sum_{t\ge k}\gamma^{t-k}r(s_t,a_t)\mid\cF_k]-V^\pi(s_k)\le H-V^*(s_k)+\Delta_\pi\le H+\Delta_\pi$,
so $\E L^2\le2(H+\Delta_\pi)^2$. Collecting terms gives at most
$2H^2+8(H+\Delta_\pi)^2+8\Delta_\pi^2\le18(H+\Delta_\pi)^2$. For the second claim apply
Lemma~\ref{lem:resolvent} to the state chain of $\pi$ with $x(s)=\pi(\Var_PV^*)(s)$, divide by
$\gamma^2$ and use $(1+\gamma)/\gamma^2\le6$ for $\gamma\ge1/2$.
\end{proof}

\begin{lemma}[Bernstein]\label{lem:bernstein}
If $Y_1,\dots,Y_N$ are i.i.d., $Y_i\in[0,H]$, with variance $\sigma^2$, then for $t>0$, with
probability at least $1-2e^{-t}$, $|\frac1N\sum_iY_i-\E Y_1|\le\sqrt{2\sigma^2t/N}+Ht/(3N)$.
\end{lemma}
This is the standard Bernstein inequality \citep[e.g.][Thm.~2.10]{boucheron2013concentration}.

%% file: app_upper.tex
\section{Proof of Theorem~\ref{thm:upper}}\label{app:upper}

Throughout, $\Ph$ is the empirical kernel built from $N$ samples per pair, $\Vh^*$, $\Qh^*$,
$\pih$ are the soft-optimal value, $Q$-function and policy of $(\Ph,r,\gamma,\beta,\pref)$, and
\begin{gather*}
f=\Qh^*-Q^*,\qquad g=\Vh^*-V^*,\qquad \xi'=\gamma(\Ph-P)\Vh^*,\\
\kappa=\beta\KL(\pih\|\pi^*),\qquad \kappa'=\beta\KL(\pi^*\|\pih),
\end{gather*}
$\Delta=\max_s(V^*(s)-V^{\pih}(s))$ and $\iota=\log(4K(N+1)/\delta)$.

\subsection{Decoupling}

For a state $s$ and $u\in\R$ let $\widehat M_{s,u}$ be the MDP that coincides with the empirical
MDP except at $s$, where every action returns to $s$ with probability one and earns reward
$(1-\gamma)u$. Let $\Vh_{s,u}$ be its soft-optimal value. This is the construction of
\citet{agarwal2020model} for unregularized MDPs.

\begin{lemma}\label{lem:absorb}
(i) $\Vh_{s,u}(s)=u$. (ii) $\Vh_{s,\Vh^*(s)}=\Vh^*$. (iii) $\|\Vh_{s,u}-\Vh_{s,u'}\|\le|u-u'|$.
(iv) $\Vh_{s,u}$ is a function of the samples at states other than $s$ only.
(v) If $u\in[0,H]$ then $0\le\Vh_{s,u}\le H$.
\end{lemma}
\begin{proof}
(i) At $s$ all actions have the same value $(1-\gamma)u+\gamma v(s)$, so the soft Bellman
equation at $s$ reads $v(s)=(1-\gamma)u+\gamma v(s)$. (ii) $\Vh^*$ satisfies the soft Bellman
equations of $\widehat M_{s,u}$ at every $s'\ne s$ (same transitions and rewards there) and at $s$
when $u=\Vh^*(s)$; uniqueness of the fixed point (Lemma~\ref{lem:basic}) gives the claim.
(iii) The two soft Bellman operators differ only at $s$, by $(1-\gamma)(u-u')$; so
$\|\Vh_{s,u}-\Vh_{s,u'}\|\le\gamma\|\Vh_{s,u}-\Vh_{s,u'}\|+(1-\gamma)|u-u'|$. (iv) is by
construction and (v) follows from Lemma~\ref{lem:basic}(iii) since all rewards lie in $[0,1]$.
\end{proof}

\begin{lemma}[decoupled noise bound]\label{lem:noise}
With probability at least $1-\delta$, for every $(s,a)\in\cZ$,
\begin{gather*}
|\xi'(s,a)|\le\sqrt{2\iota/N}\,\big(\sigma_{V^*}(s,a)+\sigma_g(s,a)\big)+3H\iota/N,\\
\xi'(s,a)^2\le\frac{6\iota}{N}\big(\sigma_{V^*}^2(s,a)+\sigma_g^2(s,a)\big)+\frac{27H^2\iota^2}{N^2},
\end{gather*}
where $\sigma_v^2(s,a)=\Var_Pv(s,a)$ for $v\in\{V^*,g\}$.
\end{lemma}
\begin{proof}
Let $U=\{0,H/N,2H/N,\dots,H\}$ ($N+1$ points). For fixed $s$, $u\in U$ and $a\in\cA_s$,
condition on the samples at states other than $s$: by Lemma~\ref{lem:absorb}(iv,v),
$\Vh_{s,u}$ is then a fixed function with values in $[0,H]$, and $\Ph(\cdot\mid s,a)$ is the
empirical law of $N$ fresh draws from $P(\cdot\mid s,a)$. Lemma~\ref{lem:bernstein} with
$t=\iota$ and a union bound over the $K(N+1)$ triples give, with probability at least $1-\delta$,
for all of them,
$|(\Ph-P)\Vh_{s,u}(s,a)|\le\sqrt{2\iota\Var_P\Vh_{s,u}(s,a)/N}+H\iota/N$.
On this event fix $(s,a)$ and let $u_s\in U$ be closest to $\Vh^*(s)\in[0,H]$. By
Lemma~\ref{lem:absorb}(ii,iii), $\|\Vh_{s,u_s}-\Vh^*\|\le H/(2N)$, hence
$|(\Ph-P)\Vh^*(s,a)|\le|(\Ph-P)\Vh_{s,u_s}(s,a)|+H/N$ and
$\sqrt{\Var_P\Vh_{s,u_s}}\le\sqrt{\Var_P\Vh^*}+H/(2N)\le\sigma_{V^*}+\sigma_g+H/(2N)$ (the standard
deviation is a seminorm bounded by the sup norm). Since $\iota\ge1$,
$\sqrt{2\iota/N}\,H/(2N)\le H\iota/N$, and $\gamma\le1$ gives the first bound; the second
follows from $(x+y+z)^2\le3(x^2+y^2+z^2)$.
\end{proof}

\subsection{Identities}

\begin{lemma}\label{lem:identities}
(i) $f=\xi'+\gamma Pg$. (ii) $g=\pih f-\kappa=\pi^*f+\kappa'$.
(iii) $V^{\pih}_{\Ph}-V^{\pih}=\Vh^*-V^{\pih}$ satisfies $d=\pih(\xi'+\gamma Pd)$.
\end{lemma}
\begin{proof}
(i) $\Qh^*-Q^*=\gamma\Ph\Vh^*-\gamma PV^*=\gamma(\Ph-P)\Vh^*+\gamma P(\Vh^*-V^*)$.
(ii) Lemma~\ref{lem:gibbs} applied to $\Vh^*$ and to $V^*$ with the same $\mu$ gives
$\Vh^*=\mu\Qh^*-\beta\KL(\mu\|\pref)+\beta\KL(\mu\|\pih)$ and
$V^*=\mu Q^*-\beta\KL(\mu\|\pref)+\beta\KL(\mu\|\pi^*)$; subtract with $\mu=\pih$ and with $\mu=\pi^*$.
(iii) $\Vh^*=V^{\pih}_{\Ph}=\pih(r+\gamma\Ph\Vh^*)-\beta\KL_{\pih}$ and
$V^{\pih}=\pih(r+\gamma PV^{\pih})-\beta\KL_{\pih}$; subtract and use (i)'s decomposition.
\end{proof}

By (ii), $g$ is the value of $\pih$ under $P$ for the reward $\xi'$ and state cost $\kappa$, and
the value of $\pi^*$ for the reward $\xi'$ and state cost $-\kappa'$; in both cases
Lemma~\ref{lem:mvi} applies with $q=f$. By (iii), $\Vh^*-V^{\pih}$ is the value of $\pih$ for
reward $\xi'$ and no cost.

\subsection{A closed system of inequalities}

Define
\begin{gather*}
X=\max_s\E^{\pih}_s\sum_t\gamma^t\xi'(s_t,a_t)^2,\qquad
X_*=\max_s\E^{\pi^*}_s\sum_t\gamma^t\xi'(s_t,a_t)^2,\\
Y=\max(X,X_*),\qquad G=\|g\|.
\end{gather*}
All statements below hold on the event of Lemma~\ref{lem:noise} and are deterministic
statements about the realized data: given the data, $\pih$, $\xi'$, $g$, $f$, $\kappa$ and $\kappa'$
are fixed functions, and $\E^{\pih}_s$, $\E^{\pi^*}_s$ denote expectations over a fresh trajectory
of the true MDP. The Markov property and Lemmas~\ref{lem:pdl}, \ref{lem:gn}, \ref{lem:resolvent}
and~\ref{lem:mvi} therefore apply for each fixed data set. Cauchy--Schwarz with weights
$\gamma^t$ gives $\E^{\pi}_s\sum_t\gamma^t|\xi'_t|\le\sqrt{HX_\pi}$ and
$\E^{\pi}_s(\sum_t\gamma^t\xi'_t)^2\le HX_\pi$ for $\pi\in\{\pih,\pi^*\}$.

\begin{lemma}\label{lem:gdelta}
$-\sqrt{HX_*}\le g\le\sqrt{HX}$, so $G\le\sqrt{HY}$; and $\Delta\le\sqrt{HX}+\sqrt{HX_*}\le2\sqrt{HY}$.
\end{lemma}
\begin{proof}
By Lemma~\ref{lem:mvi} and Lemma~\ref{lem:identities}(ii),
$g(s)=\E^{\pih}_s\sum_t\gamma^t(\xi'_t-\kappa(s_t))\le\E^{\pih}_s\sum_t\gamma^t|\xi'_t|$ and
$g(s)=\E^{\pi^*}_s\sum_t\gamma^t(\xi'_t+\kappa'(s_t))\ge-\E^{\pi^*}_s\sum_t\gamma^t|\xi'_t|$.
For $\Delta$: $V^*-V^{\pih}=-g+(\Vh^*-V^{\pih})$ and, by Lemma~\ref{lem:identities}(iii) and
Lemma~\ref{lem:mvi}, $(\Vh^*-V^{\pih})(s)=\E^{\pih}_s\sum_t\gamma^t\xi'_t\le\sqrt{HX}$.
\end{proof}

\begin{lemma}\label{lem:fnoise}
For every $s_0$,
\begin{gather*}
\sum_t\gamma^{2t}\E^{\pih}_{s_0}\big[\Var_{\pih}f(s_t)+\gamma^2\Var_Pg(s_t,a_t)\big]\le3(HX+2\Delta^2+G^2),\\
\sum_t\gamma^{2t}\E^{\pi^*}_{s_0}\big[\gamma^2\Var_Pg(s_t,a_t)\big]\le15(HX_*+G^2).
\end{gather*}
Consequently $\max_s\E^{\pih}_s\sum_t\gamma^t\sigma_g^2(s_t,a_t)\le18(HX+2\Delta^2+G^2)$ and
$\max_s\E^{\pi^*}_s\sum_t\gamma^t\sigma_g^2(s_t,a_t)\le90(HX_*+G^2)$.
\end{lemma}
\begin{proof}
Under $\pih$, Lemma~\ref{lem:mvi} (reward $\xi'$, cost $\kappa$) bounds the left side by
$\E[(\sum_t\gamma^t\xi'_t-\sum_t\gamma^t\kappa(s_t)-g(s_0))^2]\le3(HX+\E(\sum_t\gamma^t\kappa(s_t))^2+G^2)$.
By Lemma~\ref{lem:pdl}, $\E[\sum_{t\ge k}\gamma^{t-k}\kappa(s_t)\mid\cF_k]=V^*(s_k)-V^{\pih}(s_k)\le\Delta$,
so Lemma~\ref{lem:gn} gives $\E(\sum_t\gamma^t\kappa(s_t))^2\le2\Delta^2$.
Under $\pi^*$ (reward $\xi'$, cost $-\kappa'$) the same argument gives
$3(HX_*+\E(\sum_t\gamma^t\kappa'(s_t))^2+G^2)$. Here
$\E[\sum_{t\ge k}\gamma^{t-k}\kappa'(s_t)\mid\cF_k]=g(s_k)-\E[\sum_{t\ge k}\gamma^{t-k}\xi'_t\mid\cF_k]\le G+\sqrt{HX_*}$,
so Lemma~\ref{lem:gn} gives $\E(\sum_t\gamma^t\kappa'(s_t))^2\le2(G+\sqrt{HX_*})^2\le4G^2+4HX_*$.
The consequences follow from Lemma~\ref{lem:resolvent} applied to the state chains of $\pih$
and $\pi^*$ with $x(s)=\pi(\sigma_g^2)(s)$, and $(1+\gamma)/\gamma^2\le6$.
\end{proof}

\begin{lemma}\label{lem:closed}
If $N\ge c_0H\iota$ with $c_0=12528$, then $Y\le2600H^2\iota/N$. Consequently
$G\le51\sqrt{H^3\iota/N}$, $\Delta\le102\sqrt{H^3\iota/N}$ and $\|f\|\le52\sqrt{H^3\iota/N}$.
\end{lemma}
\begin{proof}
By Lemma~\ref{lem:noise}, Lemma~\ref{lem:tv} (with $\Delta_{\pih}=\Delta$ and $\Delta_{\pi^*}=0$) and
Lemma~\ref{lem:fnoise},
\begin{align*}
X&\le\tfrac{6\iota}{N}\big(108(H+\Delta)^2+18(HX+2\Delta^2+G^2)\big)+\tfrac{27H^3\iota^2}{N^2},\\
X_*&\le\tfrac{6\iota}{N}\big(108H^2+90(HX_*+G^2)\big)+\tfrac{27H^3\iota^2}{N^2}.
\end{align*}
Insert $G^2\le HY$ and $\Delta^2\le4HY$ (Lemma~\ref{lem:gdelta}) and $(H+\Delta)^2\le2H^2+2\Delta^2$:
both right sides are at most $\frac{6\iota}{N}(216H^2+1044HY)+\frac{27H^3\iota^2}{N^2}$. Hence
$Y\le1296H^2\iota/N+6264H\iota Y/N+27H^3\iota^2/N^2$. For $N\ge12528H\iota$ the coefficient of
$Y$ is at most $1/2$, and $27H^3\iota^2/N^2\le0.003H^2\iota/N$, so $Y\le2600H^2\iota/N$. The bounds
on $G$ and $\Delta$ follow from Lemma~\ref{lem:gdelta}. For $f$, Lemma~\ref{lem:identities}(i)
gives $\|f\|\le\|\xi'\|+G$, and by Lemma~\ref{lem:noise} with $\sigma_{V^*}\le H/2$,
$\sigma_g\le G$ and $N\ge c_0H\iota$, $H\ge2$: $\|\xi'\|\le\sqrt{H^2\iota/(2N)}+51\sqrt2\,\sqrt{H^3}\iota/N+3H\iota/N\le\sqrt{H^3\iota/N}$.
\end{proof}

\subsection{The fast rate}

\begin{lemma}\label{lem:fast}
If $N\ge c_0H\iota$ and $N\ge c_1H^3\iota/\beta^2$ with $c_1=1836^2$, then
$\Delta\le46800\,H^3\iota/(\beta N)$.
\end{lemma}
\begin{proof}
By Lemma~\ref{lem:closed}, $2\|f\|\le104\sqrt{H^3\iota/N}\le\beta$, so $\osc f(s,\cdot)\le\beta$
for every $s$. Since $\pih(\cdot\mid s)\propto\pi^*(\cdot\mid s)e^{f(s,\cdot)/\beta}$,
Lemma~\ref{lem:klvar} with $b=1$ gives $\kappa(s)\le\frac{3}{4\beta}\Var_{\pih}f(s)$. By
Lemma~\ref{lem:pdl}, Lemma~\ref{lem:resolvent} (state chain of $\pih$, $x=\Var_{\pih}f$) and
Lemma~\ref{lem:fnoise},
\begin{align*}
\Delta=\max_s\E^{\pih}_s\sum_t\gamma^t\kappa(s_t)&\le\frac{3(1+\gamma)}{4\beta}\max_s\E^{\pih}_s\sum_t\gamma^{2t}\Var_{\pih}f(s_t)\\
&\le\frac{9}{2\beta}(HX+2\Delta^2+G^2)\le\frac9\beta(HY+\Delta^2).
\end{align*}
Using $\Delta\le102\sqrt{H^3\iota/N}$ and $N\ge c_1H^3\iota/\beta^2$,
$\frac9\beta\Delta^2\le\frac{918\sqrt{H^3\iota/N}}{\beta}\Delta\le\frac12\Delta$. Hence
$\Delta\le\frac{18}{\beta}HY\le46800H^3\iota/(\beta N)$.
\end{proof}

\begin{proof}[Proof of Theorem~\ref{thm:upper}]
Take $c_0=12528$. On the event of Lemma~\ref{lem:noise} (probability at least $1-\delta$),
Lemma~\ref{lem:closed} gives $\|f\|\le52\sqrt{H^3\iota/N}$ and $\Delta\le102\sqrt{H^3\iota/N}$, and
Lemma~\ref{lem:fast} gives
$\Delta\le46800H^3\iota/(\beta N)$ when $N\ge c_1H^3\iota/\beta^2$. If instead
$N<c_1H^3\iota/\beta^2$, then $\sqrt{H^3\iota/N}<\sqrt{c_1}\,H^3\iota/(\beta N)$, so in all cases
$\Delta\le C\min\{\sqrt{H^3\iota/N},H^3\iota/(\beta N)\}$ with $C=102\sqrt{c_1}<1.9\cdot10^5$, and also
$\|f\|\le C\sqrt{H^3\iota/N}$.
\end{proof}

\begin{remark}[where the extra factor $H$ is avoided]
The proof above replaces the sum of per-step worst cases, $H\|f\|^2/\beta\asymp H^4\iota/(\beta N)$, by
the across-action variance of $f$ along the learner's own trajectories, whose discounted sum is controlled by the martingale identity of
Lemma~\ref{lem:mvi}: by Lemma~\ref{lem:fnoise} it is at most $3(HX+2\Delta^2+G^2)$, and $HX$
is of order $H\cdot H^2\iota/N$ because of the total variance bound of Lemma~\ref{lem:tv}.
\end{remark}

\begin{proof}[Proof of Corollary~\ref{cor:samples}]
Let $\epsilon\in(0,H]$. If $\beta\ge\epsilon$ and $N\ge CH^3\iota/(\beta\epsilon)$ then
$\Delta\le CH^3\iota/(\beta N)\le\epsilon$; if $\beta<\epsilon$ and $N\ge C^2H^3\iota/\epsilon^2$ then
$\Delta\le C\sqrt{H^3\iota/N}\le\epsilon$. Both conditions are implied by
$N\ge C^2H^3\iota/(\epsilon\max\{\epsilon,\beta\})$ since $C\ge1$; adding the burn-in
$N\ge c_0H\iota$ and multiplying by $K$ gives the claim.
\end{proof}

%% file: app_lower.tex
\section{Proof of Theorem~\ref{thm:lower}}\label{app:lower}

\paragraph{The two-arm gadget.}
Fix $\gamma\ge4/7$ (so $H\ge7/3$), $p_0=(4\gamma-1)/(3\gamma)\ge3/4$ and $0\le\Delta\le(1-p_0)/2$.
The MDP $M_\theta$, $\theta\in\{+1,-1\}$, has a decision state $x$ with two actions and three
single-action states $y_1,y_2,w$: action $i$ at $x$ moves to $y_i$ with reward $0$; at $y_i$ the
reward is $1$ and the next state is $y_i$ with probability $p_i$ and $w$ otherwise; $w$ is
absorbing with reward $0$; $\pref(\cdot\mid x)$ is uniform; $p_1=p_0+\theta\Delta$ and
$p_2=p_0-\theta\Delta$. There are $K=5$ pairs, and only the samples at $y_1,y_2$ are random.
Single-action states contribute no KL term, so by Lemma~\ref{lem:pdl}, for every policy $\pi$,
\begin{equation}\label{eq:lb-loss}
\begin{gathered}
V^*_\theta(x)-V^\pi_\theta(x)=\beta\KL\big(\pi(\cdot\mid x)\,\|\,\pi^*_\theta(\cdot\mid x)\big),\\
\pi^*_\theta(1\mid x)=\sigma(\theta u),\qquad u=\frac{Q(p_0+\Delta)-Q(p_0-\Delta)}{\beta},
\end{gathered}
\end{equation}
where $Q(p)=\gamma/(1-\gamma p)$ is the value of action $i$ when $p_i=p$ and $\sigma$ is the logistic
function. Since $1-\gamma p_0=\frac43(1-\gamma)=\frac{4}{3H}$,
\begin{equation}\label{eq:lb-gap}
\beta u=\frac{2\gamma^2\Delta}{(1-\gamma p_0)^2-\gamma^2\Delta^2}\ \ge\ \frac{2\gamma^2\Delta}{(1-\gamma p_0)^2}=\frac98\gamma^2H^2\Delta .
\end{equation}

\begin{lemma}[two-point loss]\label{lem:twopoint}
For $u>0$ and $q\in[0,1]$,
$\KL(\mathrm{Ber}(q)\|\mathrm{Ber}(\sigma(u)))+\KL(\mathrm{Ber}(q)\|\mathrm{Ber}(\sigma(-u)))\ge2\log\cosh(u/2)$,
and $\log\cosh v\ge0.3\min\{v^2,v\}$ for $v\ge0$.
\end{lemma}
\begin{proof}
The left side is convex in $q$ and symmetric under $q\mapsto1-q$, so it is minimized at $q=1/2$,
where it equals $-\log(4\sigma(u)\sigma(-u))=2\log\cosh(u/2)$. For $v\le1$,
$\log\cosh v\ge v^2/2-v^4/12\ge\frac5{12}v^2$; for $v\ge1$, $\log\cosh v\ge v-\log2\ge0.3v$
(at $v=1$, $\log\cosh 1=0.434$, and $v-\log 2-0.3v$ is increasing).
\end{proof}

\begin{lemma}[information]\label{lem:lb-info}
Let $\mathrm{kl}(a,b)$ denote the Bernoulli KL divergence. For $b\in\{p_0\pm\Delta\}$ and
$a\in\{p_0\pm\Delta\}$, $\mathrm{kl}(a,b)\le16\Delta^2/(1-p_0)=48\gamma H\Delta^2$.
\end{lemma}
\begin{proof}
$\mathrm{kl}(a,b)\le(a-b)^2/(b(1-b))$, $|a-b|\le2\Delta$, and $b(1-b)\ge\frac12\cdot\frac{1-p_0}{2}$
because $b\ge p_0-\Delta\ge1/2$ and $1-b\ge(1-p_0)/2$. Finally $1-p_0=(1-\gamma)/(3\gamma)$.
\end{proof}

\begin{proof}[Proof of Theorem~\ref{thm:lower}(a)]
Let $\widehat\pi$ be any estimator and write $q=\widehat\pi(1\mid x)$. By \eqref{eq:lb-loss} and
Lemma~\ref{lem:twopoint}, for every realization of the data,
$L_+(q)+L_-(q)\ge2\beta\log\cosh(u/2)$, where $L_\theta$ is the loss under $M_\theta$. With
$\mathbb P_\theta$ the law of the data,
$\max_\theta\E_\theta L_\theta\ge\frac12\int(L_+d\mathbb P_++L_-d\mathbb P_-)\ge\beta\log\cosh(u/2)\,(1-\mathrm{TV}(\mathbb P_+,\mathbb P_-))$.
The data are $N$ draws at $y_1$ and $N$ at $y_2$, so by Lemma~\ref{lem:lb-info}
$\KL(\mathbb P_+\|\mathbb P_-)\le2N\cdot48\gamma H\Delta^2$. Choose $\Delta=(768\gamma HN)^{-1/2}$,
which satisfies $\Delta\le(1-p_0)/2=(6\gamma H)^{-1}$ whenever $N\ge3\gamma H/64$, in particular when
$N\ge H/16$. Then $\KL\le1/8$, $\mathrm{TV}\le\sqrt{\KL/2}\le1/4$, and by \eqref{eq:lb-gap}
$\beta u\ge\frac{9}{8\sqrt{768}}\gamma^{3/2}\sqrt{H^3/N}\ge g_0:=0.0175\sqrt{H^3/N}$ (using
$\gamma^{3/2}\ge(4/7)^{3/2}$). By Lemma~\ref{lem:twopoint} and monotonicity of $\log\cosh$,
\[
\max_\theta\E_\theta\big[V^*_\theta(x)-V^{\widehat\pi}_\theta(x)\big]\ge\tfrac34\beta\cdot0.3\min\Big\{\frac{g_0^2}{4\beta^2},\frac{g_0}{2\beta}\Big\}
\ge1.7\cdot10^{-5}\min\Big\{\frac{H^3}{\beta N},\sqrt{\frac{H^3}{N}}\Big\}.\qedhere
\]
\end{proof}

\begin{proof}[Proof of Theorem~\ref{thm:lower}(b)]
Take $m$ independent copies of the gadget (decision states $x_j$, loop states $y_{j1},y_{j2}$)
sharing one absorbing state $w$, so $K=4m+1$, with $\rho$ uniform on $\{x_j\}$ and parameter
$\theta\in\{\pm1\}^m$. By Lemma~\ref{lem:pdl},
$V^*_\theta(\rho)-V^\pi_\theta(\rho)=\frac1m\sum_j\beta\KL(\pi(\cdot\mid x_j)\|\pi^*_{\theta_j}(\cdot\mid x_j))$.
Consider any algorithm that makes $T$ adaptively chosen queries (possibly randomized) and then
outputs $\widehat\pi$. Let $n_j$ be the number of queries at $y_{j1}$ or $y_{j2}$; only these
observations depend on $\theta_j$. For $\theta$ and $j$ let $\theta^{j\pm}$ be $\theta$ with
coordinate $j$ set to $\pm1$, and $\mathbb P_\theta$ the law of the transcript. The divergence
decomposition for adaptive sampling \citep[Lemma~15.1]{lattimore2020bandit} and
Lemma~\ref{lem:lb-info} give
$\KL(\mathbb P_{\theta^{j+}}\|\mathbb P_{\theta^{j-}})\le48\gamma H\Delta^2\,\E_{\theta^{j+}}n_j$.
Averaging over $\theta$ uniform (Assouad's argument) and using the two-point bound for each
coordinate,
\[
\bar{\E}\big[V^*_\theta(\rho)-V^{\widehat\pi}_\theta(\rho)\big]\ge\frac{\beta\log\cosh(u/2)}{m}\sum_j\overline{\Big(1-\sqrt{24\gamma H\Delta^2\,\E_{\theta^{j+}}n_j}\Big)},
\]
where the bar denotes averaging over $\theta_{-j}$ (respectively $\theta$). By concavity of the
square root and $\overline{\E_{\theta^{j+}}n_j}\le2\,\bar\E[n_j]$, $\sum_j\bar\E n_j\le T$, the right side
is at least $\beta\log\cosh(u/2)(1-\sqrt{48\gamma H\Delta^2T/m})$, which is at least
$\frac12\beta\log\cosh(u/2)$ when $T\le m/(192\gamma H\Delta^2)$.
Now fix $\epsilon>0$ and choose $\Delta$ so that $\frac12\beta\cdot0.3\min\{u^2/4,u/2\}=2\epsilon$ with
$\beta u=\frac98\gamma^2H^2\Delta$ (a lower bound on the true gap by \eqref{eq:lb-gap}, so the loss is
then at least $2\epsilon>\epsilon$). This gives
$\Delta^2\le\frac{712\,\epsilon\max\{\epsilon,\beta\}}{(9/8)^2\gamma^4H^4}$ (in the branch $u\le2$ the solution is $\Delta^2=\frac{160}{3}\frac{\beta\epsilon}{(9/8)^2\gamma^4H^4}$, in the branch $u\ge2$ it is $\Delta^2=(\frac{80}{3})^2\frac{\epsilon^2}{(9/8)^2\gamma^4H^4}$),
so any algorithm with $\sup_\theta\E_\theta[V^*_\theta(\rho)-V^{\widehat\pi}_\theta(\rho)]\le\epsilon$ must have
\[
T>\frac{m}{192\gamma H\Delta^2}\ \ge\ c'\,\frac{m\,H^3}{\epsilon\max\{\epsilon,\beta\}},\qquad c'=\frac{(9/8)^2\gamma^3}{192\cdot 712}\ge 1.7\cdot10^{-6}.
\]
The constraint $\Delta\le(6\gamma H)^{-1}$ holds when $\epsilon\max\{\epsilon,\beta\}\le1.6\cdot10^{-5}H^2$
(use $\gamma^2\ge(4/7)^2$). Finally, a family with at most $K\ge5$ pairs allows
$m=\lfloor(K-1)/4\rfloor\ge(K-4)/4\ge K/20$ gadgets, which gives
$T\ge8\cdot10^{-8}KH^3/(\epsilon\max\{\epsilon,\beta\})$. Both parts of Theorem~\ref{thm:lower} therefore hold
with $c=8\cdot10^{-8}$.
\end{proof}

\paragraph{The trivial regime.} We prove Proposition~\ref{prop:trivial}; it shows that Theorem~\ref{thm:lower}(b) cannot hold for $\epsilon\ge H^3/(8\beta)$.
\begin{proof}[Proof of Proposition~\ref{prop:trivial}]
By Lemma~\ref{lem:pdl}, $V^*-V^{\pref}=\E^{\pref}\sum_t\gamma^t\beta\KL(\pref\|\pi^*)(s_t)$ and
$\KL(\pref\|\pi^*)(s)=\log\E_{\pref}e^{(Q^*(s,\cdot)-\pref Q^*(s))/\beta}\le\osc(Q^*)^2/(8\beta^2)\le H^2/(8\beta^2)$
by Hoeffding's lemma and $Q^*\in[0,H]$.
\end{proof}

%% file: app_asymp.tex
\section{Proofs of Theorems~\ref{thm:asymp} and~\ref{thm:design}}\label{app:asymp}

Fix the instance. For $z\in\cZ$ let $\mathrm{supp}_z=\{s':P(s'\mid z)>0\}$; the empirical kernel
is supported on these sets. Let $\mathcal R$ be the set of states reachable from $\rho$; since
$\pref$ has full support, every policy only visits $\mathcal R$, and
$d(s):=\sum_{t\ge0}\gamma^t\Pr^{\pi^*}_\rho(s_t=s)>0$ (unnormalized, total mass $H$) for
$s\in\mathcal R$. Write $P_{\pi^*}$ for the $K\times K$ matrix
$P_{\pi^*}[(s,a),(s',a')]=P(s'\mid s,a)\pi^*(a'\mid s')$, $M=(I-\gamma P_{\pi^*})^{-1}$,
$\sigma_z^2=\Var_PV^*(z)$, $F_s=\mathrm{diag}(\pi^*_s)-\pi^*_s\pi_s^{*\top}$ and
$W=\mathrm{blockdiag}_s(d(s)F_s)$. For $q\in\R^{\cZ}$ let
$\Phi(q)=V^*(\rho)-V^{\pi_q}(\rho)$.

\begin{lemma}[smoothness]\label{lem:smooth}
(i) On a neighbourhood of $P$ within the kernels supported on $\{\mathrm{supp}_z\}$, the map
$P'\mapsto Q^*_{P'}$ is $C^\infty$, with derivative $h\mapsto M\gamma hV^*$ at $P$.
(ii) $\Phi$ is $C^\infty$ near $Q^*$, $\Phi\ge0$, $\Phi(Q^*)=0$, $\nabla\Phi(Q^*)=0$ and
$\nabla^2\Phi(Q^*)=W/\beta$.
\end{lemma}
\begin{proof}
(i) $Q^*_{P'}$ solves $F(q,P')=q-r-\gamma P'\,\mathrm{soft}(q)=0$ with
$\mathrm{soft}(q)(s)=\beta\log\sum_a\pref(a\mid s)e^{q(s,a)/\beta}$. $F$ is $C^\infty$ and
$\partial_qF(Q^*,P)=I-\gamma P_{\pi^*}$, whose inverse is $M$ (the Jacobian of $\mathrm{soft}$ at
$q$ has rows $\pi_q(\cdot\mid s)$). The implicit function theorem gives (i).
(ii) By Lemma~\ref{lem:pdl}, $\Phi(q)=\beta\sum_sd^{\pi_q}_\rho(s)\KL(\pi_q(\cdot\mid s)\|\pi^*(\cdot\mid s))$.
The occupancy $d^{\pi}_\rho$ is a smooth function of $\pi$, and for the exponential family
$\pi_q(\cdot\mid s)$ with natural parameter $q(s,\cdot)/\beta$,
$\KL(\pi_q\|\pi^*)(s)=\frac{1}{2\beta^2}(q-Q^*)_s^\top F_s(q-Q^*)_s+O(\|q-Q^*\|^3)$
(the KL is the Bregman divergence of the log-partition function, whose Hessian is the
covariance $F_s$).
\end{proof}

\begin{proof}[Proof of Theorem~\ref{thm:asymp}(i)]
By the multivariate central limit theorem, $\sqrt N(\Ph(\cdot\mid z)-P(\cdot\mid z))$ converge jointly
to independent Gaussian vectors $G_z$ with $\mathrm{Cov}(G_z^\top V^*)=\sigma_z^2$. By
Lemma~\ref{lem:smooth}(i) and the delta method, $\sqrt N(\Qh^*-Q^*)\to Z:=\gamma M(G_z^\top V^*)_z$
in distribution, so $Z\sim\mathcal N(0,\Sigma_Q)$ with $\Sigma_Q=\gamma^2M\,\mathrm{diag}(\sigma^2)M^\top$.
By Lemma~\ref{lem:smooth}(ii), $N\Phi(\Qh^*)=\frac1{2\beta}(\sqrt N(\Qh^*-Q^*))^\top W(\sqrt N(\Qh^*-Q^*))+N\,O(\|\Qh^*-Q^*\|^3)$
on events of probability tending to one, and the remainder is $O_p(N^{-1/2})$. The continuous
mapping theorem gives $N\Phi(\Qh^*)\to\frac1{2\beta}Z^\top WZ$.
\end{proof}

\begin{proof}[Proof of Theorem~\ref{thm:asymp}(ii)]
It suffices to show that $N\Phi(\Qh^*)$ is uniformly integrable, since
$\E\frac1{2\beta}Z^\top WZ=\frac1{2\beta}\mathrm{tr}(W\Sigma_Q)$. Let $\xi=\gamma(\Ph-P)V^*$ and
$M_N=\|\xi\|$. With $f=\Qh^*-Q^*$ and $g=\Vh^*-V^*$, the identities of
Lemma~\ref{lem:identities} also hold with $(\Ph,\xi)$ in place of $(P,\xi')$:
$f=\xi+\gamma\Ph g$ and $\pi^*f\le g\le\pih f$. Hence
$(I-\gamma\Ph_{\pi^*})^{-1}\xi\le f\le(I-\gamma\Ph_{\pih})^{-1}\xi$ and $\|f\|\le HM_N$.
Choose $r_0>0$ such that $\Phi(q)\le C_0\|q-Q^*\|^2$ for $\|q-Q^*\|\le r_0$
(Lemma~\ref{lem:smooth}(ii)). On $A_N=\{HM_N\le r_0\}$, $N\Phi(\Qh^*)\le C_0H^2NM_N^2$, and by
Hoeffding's inequality $\Pr(\sqrt NM_N>t)\le2Ke^{-2t^2/H^2}$, so $\sup_N\E[(NM_N^2)^2]<\infty$.
On $A_N^c$, $\Phi(\Qh^*)\le H+H^2$ because $\KL(\pih\|\pref)\le\osc(\Qh^*)/\beta\le H/\beta$ gives
$V^{\pih}\ge-H^2$; and $\Pr(A_N^c)\le2Ke^{-2Nr_0^2/H^4}$, so $N(H+H^2)\Pr(A_N^c)\to0$.
\end{proof}

\subsection{Local minimax lower bound over adaptive designs}

Parametrize each $P(\cdot\mid z)$ by its coordinates on $\mathrm{supp}_z$ minus one reference
state; this gives $\theta\in\Theta\subset\R^D$, an open set, with the true $\theta_0$. An
\emph{algorithm} queries pairs $z_1,\dots,z_T$, each chosen as a measurable function of the past
observations and independent randomness, observes $s'_k\sim P_\theta(\cdot\mid z_k)$, and outputs a
policy $\widetilde\pi$. Let $n_z$ be the number of queries at $z$ and $I_z(\theta)$ the Fisher
information of one draw from $P_\theta(\cdot\mid z)$ (a block supported on the coordinates of
$z$). The score of the transcript is a sum of martingale differences, so its Fisher information
is $\mathcal I_T(\theta)=\sum_z\E_\theta[n_z]I_z(\theta)$.

\begin{lemma}[van Trees inequality; \citealt{gill1995applications}]\label{lem:vantrees}
Let $\lambda$ be a $C^1$ prior density on a compact ball $B\subset\Theta$, vanishing on $\partial B$,
with finite Fisher information matrix $J(\lambda)=\int\nabla\lambda\nabla\lambda^\top/\lambda$. For
$\varphi\in C^1(\Theta)$, any estimator $\widehat\varphi$ and any $b\in\R^D$,
$\int\E_\theta(\widehat\varphi-\varphi(\theta))^2\lambda(\theta)d\theta\ge\frac{(\int b^\top\nabla\varphi\,\lambda)^2}{\int b^\top\mathcal I_T(\theta)b\,\lambda+b^\top J(\lambda)b}$.
\end{lemma}

Let $\psi(\theta)=Q^*_{P_\theta}$, $\dot\psi=\nabla\psi(\theta_0)$ (by Lemma~\ref{lem:smooth}(i),
column block $z$ is $\gamma M e_z\,\nabla_{\theta_z}(P_\theta(\cdot\mid z)^\top V^*)$), and
$\|e\|_W^2=e^\top W e$, a seminorm that vanishes on per-state constants and on states outside
$\mathcal R$ (where $d=0$; the same holds for $W_\theta$ in the fixed-support neighbourhood). For a policy
$\pi$ with no zero on $\mathcal R$, let $\ell(\pi)(s,a)=\beta\log(\pi(a\mid s)/\pref(a\mid s))$ for $s\in\mathcal R$ and
$\ell(\pi)(s,a)=\psi(\theta_0)(s,a)$ for $s\notin\mathcal R$, a finite vector; if $\pi$ has a zero on $\mathcal R$,
$\ell(\pi)$ is not defined and we set $\|\ell(\pi)-q\|_W=\infty$ for every $q$. Write $\Phi_\theta$, $W_\theta$ for
the quantities of the instance $\theta$.

\begin{lemma}[reduction to a quadratic loss]\label{lem:reduce}
For every $\eta\in(0,1)$ there are a neighbourhood $\mathcal N$ of $\theta_0$ and a map
$\pi\mapsto\widehat\psi(\pi)\in\R^{\cZ}$ such that
$\Phi_\theta(\pi)\ge(1-\eta)\frac1{2\beta}\|\widehat\psi(\pi)-\psi(\theta)\|_W^2$ for all $\theta\in\mathcal N$ and all
policies $\pi$.
\end{lemma}
\begin{proof}
\emph{Local part.} By Lemma~\ref{lem:smooth}(ii) applied to every instance near $\theta_0$
(the Hessian $W_\theta/\beta$ and the third derivatives are continuous in $\theta$, and
$W_\theta\to W$), there are $r_0>0$ and a neighbourhood $\mathcal N_0$ such that
$\Phi_\theta(\pi)\ge(1-\eta)\frac1{2\beta}\|\ell(\pi)-\psi(\theta)\|_W^2$ whenever $\theta\in\mathcal N_0$ and
$\|\ell(\pi)-\psi(\theta)\|_W\le3r_0$; here we use that $\|\cdot\|_W$ is a norm on per-state
centred vectors over $\mathcal R$, so a small $\|\ell(\pi)-\psi(\theta)\|_W$ forces $\pi$ to be
close to $\pi^*_\theta$ on $\mathcal R$.
\emph{Far part.} $\Phi_\theta(\pi)$ is jointly continuous in $(\theta,\pi)$ on the compact set
$\overline{\mathcal N_0}\times\{\text{policies}\}$ (the KL to $\pref$ is continuous because $\pref>0$).
If $\Phi_\theta(\pi)=0$ then $\pi=\pi^*_\theta$ on every state with $d^\pi_\rho>0$; since
$\pi^*_\theta$ has full support, induction along shortest paths from the support of $\rho$
shows $d^\pi_\rho>0$ on all of $\mathcal R$, so $\pi=\pi^*_\theta$ on $\mathcal R$. Hence
$c_{\rm far}=\inf\{\Phi_\theta(\pi):\theta\in\overline{\mathcal N_0},\ \|\ell(\pi)-\psi(\theta)\|_W\ge r_0\}>0$.
\emph{Construction.} Let $\mathcal N\subset\mathcal N_0$ be so small that
$\epsilon_0:=\sup_{\theta\in\mathcal N}\|\psi(\theta)-\psi(\theta_0)\|_W\le\min\{r_0,\sqrt{2\beta c_{\rm far}}\}$.
Set $\widehat\psi(\pi)=\ell(\pi)$ if $\pi$ has no zero on $\mathcal R$ and $\|\ell(\pi)-\psi(\theta_0)\|_W\le2r_0$, and
$\widehat\psi(\pi)=\psi(\theta_0)$ otherwise; in both cases $\widehat\psi(\pi)$ is a finite vector. In the first case $\|\ell(\pi)-\psi(\theta)\|_W\le3r_0$ and the local part applies. In
the second, $\|\ell(\pi)-\psi(\theta)\|_W>r_0$, so
$\Phi_\theta(\pi)\ge c_{\rm far}\ge\frac{1}{2\beta}\epsilon_0^2\ge\frac1{2\beta}\|\widehat\psi(\pi)-\psi(\theta)\|_W^2$.
\end{proof}

\begin{proof}[Proof of Theorem~\ref{thm:design}(i)]
Fix $\eta$ and take $\mathcal N$ from Lemma~\ref{lem:reduce}. Let $\lambda(u)\propto(1-\|u\|_2^2)^3$ on the unit
ball of $\R^D$; it is $C^1$, vanishes on the boundary and has $J(\lambda)=jI$ with $0<j<\infty$. For $c>0$ let
$\lambda_T(\theta)=(\sqrt T/c)^D\lambda(\sqrt T(\theta-\theta_0)/c)$, supported on the ball of radius $c/\sqrt T$ around
$\theta_0$, which lies in $\mathcal N$ for large $T$; $J(\lambda_T)=(T/c^2)J(\lambda)$. Let $W=\sum_k\omega_kv_kv_k^\top$ be an
eigendecomposition ($\omega_k\ge0$) and $\varphi_k=v_k^\top\psi$. For any algorithm, Lemma~\ref{lem:reduce} gives
\[
\sup_{\theta\in\mathrm{supp}\lambda_T}\E_\theta\Phi_\theta(\widetilde\pi)\ge\int\E_\theta\Phi_\theta(\widetilde\pi)\lambda_T
\ge\frac{1-\eta}{2\beta}\sum_k\omega_k\int\E_\theta(v_k^\top\widehat\psi(\widetilde\pi)-\varphi_k(\theta))^2\lambda_T .
\]
Let $\bar n_z=\int\E_\theta[n_z]\lambda_T$ and $\bar w_z=\bar n_z/T$, so $\bar w\in\Delta(\cZ)$, and let
$\mathbb J_c(w)=\sum_zw_zI_z(\theta_0)+J(\lambda)/c^2$. The functions $I_z$ and $\nabla\varphi_k$ are uniformly continuous
near $\theta_0$ and the support of $\lambda_T$ shrinks to $\theta_0$. Hence, with
$\epsilon_T=\max_z\sup_{\theta\in\mathrm{supp}\lambda_T}\|I_z(\theta)-I_z(\theta_0)\|\to0$ and because $\sum_zn_z=T$,
\[
B_T:=\frac1T\int\mathcal I_T\lambda_T+\frac{J(\lambda)}{c^2}=\mathbb J_c(\bar w)+E_T,\qquad\|E_T\|\le\epsilon_T,
\]
for every algorithm, and $a_{T,k}:=\int\nabla\varphi_k\lambda_T=a_k+o(1)$ with $a_k=\nabla\varphi_k(\theta_0)$ (norms are
spectral and Euclidean). Both $B_T$ and $\mathbb J_c(\bar w)$ are at least $(j/c^2)I$, so
$\|B_T^{-1}-\mathbb J_c(\bar w)^{-1}\|\le(c^2/j)^2\epsilon_T$. The denominator in Lemma~\ref{lem:vantrees} is $T\,b^\top B_Tb$;
taking $b=B_T^{-1}a_{T,k}$ (any $b$ if $a_{T,k}=0$) gives
$T\int\E_\theta(v_k^\top\widehat\psi(\widetilde\pi)-\varphi_k(\theta))^2\lambda_T\ge a_{T,k}^\top B_T^{-1}a_{T,k}\ge a_k^\top\mathbb J_c(\bar w)^{-1}a_k-\delta_T$
with $\delta_T\to0$ independent of the algorithm. Summing with the weights $\omega_k$, and using
$\sum_k\omega_ka_k^\top\mathbb J^{-1}a_k=\mathrm{tr}(W\dot\psi\,\mathbb J^{-1}\dot\psi^\top)$,
\begin{equation}\label{eq:vantrees}
\begin{split}
T\sup_{\theta\in\mathrm{supp}\lambda_T}\E_\theta\Phi_\theta(\widetilde\pi)
&\ge\frac{1-\eta}{2\beta}\Big(\mathrm{tr}\big(W\dot\psi\,\mathbb J_c(\bar w)^{-1}\dot\psi^\top\big)-\delta'_T\Big)\\
&\ge\frac{1-\eta}{2\beta}\Big(\inf_{w\in\Delta(\cZ)}\mathrm{tr}\big(W\dot\psi\,\mathbb J_c(w)^{-1}\dot\psi^\top\big)-\delta'_T\Big),
\end{split}
\end{equation}
where $\delta'_T\to0$ uniformly over algorithms. The bound has no division, so it also covers pairs and
directions with zero gradient.
As $c\to\infty$ the integrand increases, for every $w$, to $\sum_zc_z/w_z$ with the convention
$c_z/0=+\infty$ when $c_z>0$ and $0/0=0$: on blocks with $w_z>0$ the matrix
$\sum_{z'}w_{z'}I_{z'}$ is invertible (block-diagonal structure), a block with $w_z=0$ and $c_z>0$
is regularized only by $J(\lambda)/c^2$ and its contribution diverges, and a block with
$c_z=0$ has $W^{1/2}\dot\psi_z=0$ and contributes nothing. Here
$c_z=\mathrm{tr}(W\dot\psi_zI_z^{-1}\dot\psi_z^\top)=\gamma^2\sigma_z^2[M^\top WM]_{zz}$ because the
Cram\'er--Rao bound for the linear functional $P(\cdot\mid z)^\top V^*$ of one multinomial draw is
$\sigma_z^2$. The infimum of an increasing family of lower semicontinuous functions on the compact
simplex converges to the infimum of the limit, and $\inf_w\sum_zc_z/w_z=(\sum_z\sqrt{c_z})^2$ by
Cauchy--Schwarz. Let $T\to\infty$, then $c\to\infty$, then $\eta\to0$.
\end{proof}

\begin{proof}[Proof of Theorem~\ref{thm:design}(ii) and Theorem~\ref{thm:asymp}(iii)]
For a fixed allocation with $n_z=w_zT$ samples at $z$ and all $w_z>0$, the proofs of
Theorem~\ref{thm:asymp}(i,ii) go through with $\sigma_z^2$ replaced by $\sigma_z^2/w_z$ and $N$ by
$T$, so $T\,\E\Phi(\Qh^*)\to\frac1{2\beta}\sum_zc_z/w_z$. With $w_z\propto\sqrt{c_z}$ (or, if some
$c_z=0$, a small positive floor on those coordinates) this is within any factor $1+\eta$ of the
lower bound. The uniform design $w_z=1/K$, $T=KN$ gives $N\,\E\Phi\to\frac1{2\beta}\sum_zc_z=\frac1{2\beta}\mathrm{tr}(W\Sigma_Q)$,
and the first inequality in \eqref{eq:vantrees}, applied to this fixed design ($\bar w_z=1/K$ for every $T$) and
followed by the same limits (with $c$ rescaled by $\sqrt K$), shows that no estimator using $N$ samples per
pair has a smaller local asymptotic minimax risk.
\end{proof}

\input{app_uniform.tex}

\subsection{Instance dependence}

We prove Proposition~\ref{prop:revisit}.
\begin{proof}[Proof of Proposition~\ref{prop:revisit}]
Let $\zeta\sim\mathcal N(0,\gamma^2\mathrm{diag}(\sigma^2))$ and $Z=M\zeta$, so $Z\sim\mathcal N(0,\Sigma_Q)$
and $Z=\zeta+\gamma P\pi^*Z$: $Z$ is the $Q$-function of reward $\zeta$ under $\pi^*$, with value
$\bar z=\pi^*Z$. Then
$\frac1{2\beta}\mathrm{tr}(W\Sigma_Q)=\frac1{2\beta}\E^{\pi^*}_\rho\sum_t\gamma^t x(s_t)$ with
$x(s)=\E_\zeta\Var_{\pi^*}Z(s)$. By Lemma~\ref{lem:resolvent} this is at most
$\frac{1+\gamma}{2\beta}\max_{s_0}\E^{\pi^*}_{s_0}\sum_t\gamma^{2t}x(s_t)$. For fixed $\zeta$,
Lemma~\ref{lem:mvi} gives
$\sum_t\gamma^{2t}\E\Var_{\pi^*}Z(s_t)\le\E(\sum_t\gamma^t\zeta(s_t,a_t)-\bar z(s_0))^2\le\E(\sum_t\gamma^t\zeta(s_t,a_t))^2$
(the middle term is a variance around the mean $\bar z(s_0)$). Taking $\E_\zeta$ and using
independence across pairs gives $\sum_z\gamma^2\sigma_z^2\E[n_z^2]$. For the bounds, $n_z\le H$
gives $\sum_z\sigma_z^2\E n_z^2\le H\,\E^{\pi^*}\sum_t\gamma^t\Var_PV^*(s_t,a_t)\le108H^3$ by
Lemma~\ref{lem:tv}; if relevant pairs are visited at most once, $n_z^2\le n_z$ and the factor $H$
disappears.
\end{proof}

%% file: app_uniform.tex
\subsection{Uniformity of Theorem~\ref{thm:asymp}(ii) and local asymptotic minimaxity}\label{app:uniform}

\begin{proposition}\label{prop:uniform}
There are a neighbourhood $U$ of $\theta_0$ and a constant $C_U$ such that for all $N\ge1$,
\[
\sup_{\theta\in U}\Big|N\,\E_\theta\big[V^*_\theta(\rho)-V^{\pih}_\theta(\rho)\big]-\mathcal C(\theta)\Big|\le\frac{C_U}{\sqrt N},
\qquad \mathcal C(\theta)=\frac1{2\beta}\mathrm{tr}\big(W_\theta\Sigma_Q(\theta)\big).
\]
Consequently $\lim_{c\to\infty}\limsup_N N\sup_{\|\theta-\theta_0\|\le c/\sqrt N}\E_\theta[V^*_\theta(\rho)-V^{\pih}_\theta(\rho)]=\mathcal C(\theta_0)$,
and with Theorem~\ref{thm:asymp}(iii) the plug-in estimator is locally asymptotically minimax
among estimators that use $N$ samples per pair. The same holds for any fixed allocation
$w$ with all $w_z>0$ (Theorem~\ref{thm:design}(ii)), with $\mathcal C$ replaced by
$\frac1{2\beta}\sum_zc_z/w_z$ and the comparison restricted to estimators that use this allocation.
\end{proposition}

\begin{proof}
On vectors and kernels, $\|\cdot\|$ is the maximum norm (largest entry in absolute value); for a
matrix $W$, $\|W\|$ is its $\ell_\infty\to\ell_1$ norm, so that $|x^\top Wy|\le\|W\|\,\|x\|\,\|y\|$. Let
$U\subset\Theta$ be a compact neighbourhood of $\theta_0$. For $\theta\in U$ the kernel $P_\theta$ has the supports
$\mathrm{supp}_z$, and every empirical kernel drawn from it is supported on these sets, possibly with a
smaller support (so it need not lie in $\Theta$). Write $\mathcal P$ for the set of all transition kernels
supported on these sets, including those with smaller support; it is compact and convex and contains
every $P_\theta$, $\theta\in U$, and every empirical kernel.

\emph{Step 1: uniform smoothness.} The soft-optimal $Q$-function $P'\mapsto Q^*_{P'}$ is $C^\infty$ on a
neighbourhood of $\mathcal P$: it solves $F(q,P')=0$ with $F$ analytic (as in Lemma~\ref{lem:smooth}) and
$\partial_qF=I-\gamma P'_{\pi_q}$, with $P'_{\pi_q}$ the $K\times K$ state-action matrix of
Section~\ref{sec:results}. For $P'\in\mathcal P$ the inverse of $\partial_qF$ has norm at most $H$, and at most
$2H$ on a small enough neighbourhood of $\mathcal P$, so the implicit function theorem applies. Likewise
$(P',q)\mapsto\Phi_{P'}(q)=V^*_{P'}(\rho)-V^{\pi_q}_{P'}(\rho)$ is $C^\infty$, because
$V^\pi_{P'}=(I-\gamma\bar P'_\pi)^{-1}(r_\pi-\beta\KL_\pi)$, with $\bar P'_\pi(s,s')=\sum_a\pi(a\mid s)P'(s'\mid s,a)$ the
state transition matrix, is analytic in $(P',\pi)$ for $\pi$ with full support, and $\pi_q$ is analytic
in $q$. Derivatives of every order are bounded on compact sets, and $\mathcal P$ and the balls
$\{q:\|q-Q^*_\theta\|\le r_0\}$ are convex, so Taylor's theorem along segments gives constants
$L_1,L_2,L_3,r_0>0$ with the following properties. With $D_{P'}$ the derivative of $Q^*_{P'}$, for all
$P',P''\in\mathcal P$: $\|D_{P'}\|\le L_1$, $\|Q^*_{P''}-Q^*_{P'}\|\le L_1\|P''-P'\|$ and
$\|Q^*_{P''}-Q^*_{P'}-D_{P'}(P''-P')\|\le L_2\|P''-P'\|^2$. For all $\theta\in U$ and $\|q-Q^*_\theta\|\le r_0$:
$\big|\Phi_\theta(q)-\frac1{2\beta}(q-Q^*_\theta)^\top W_\theta(q-Q^*_\theta)\big|\le L_3\|q-Q^*_\theta\|^3$; here we used
$\Phi_\theta(Q^*_\theta)=0$, $\nabla\Phi_\theta(Q^*_\theta)=0$ (Lemma~\ref{lem:pdl}) and $\nabla^2\Phi_\theta(Q^*_\theta)=W_\theta/\beta$
(Lemma~\ref{lem:smooth}), for every $\theta$. Finally $\sup_{\theta\in U}\|W_\theta\|<\infty$.

\emph{Step 2: moments.} Let $\Delta_N=\sqrt N(\Ph-P_\theta)$. Its blocks are normalized sums of
bounded, centred, independent vectors, so $\sup_{\theta,N}\E_\theta\|\Delta_N\|^4<\infty$, and by Hoeffding's
inequality $\Pr_\theta(\|\Ph-P_\theta\|>t)\le2K|\cS|e^{-2Nt^2}$.

\emph{Step 3: expansion.} Let $Z_N=\sqrt N(\Qh^*-Q^*_\theta)$ and $L_N=D_{P_\theta}\Delta_N$. For every $N$,
$\E_\theta[L_NL_N^\top]=\Sigma_Q(\theta)$ exactly, because $\mathrm{Cov}(\Delta_N)$ is the multinomial covariance
and $D_{P_\theta}$ is linear. By Step 1, $\|Z_N\|,\|L_N\|\le L_1\|\Delta_N\|$ and $\|Z_N-L_N\|\le L_2\|\Delta_N\|^2/\sqrt N$. Hence
\begin{align*}
\big|\E_\theta[Z_N^\top W_\theta Z_N]-\mathrm{tr}(W_\theta\Sigma_Q(\theta))\big|
&\le\|W_\theta\|\,\E_\theta\big[\|Z_N-L_N\|(\|Z_N\|+\|L_N\|)\big]\\
&\le\frac{2L_1L_2\|W_\theta\|\,\E_\theta\|\Delta_N\|^3}{\sqrt N}\le\frac{c_1}{\sqrt N}.
\end{align*}
On $A_N=\{\|\Qh^*-Q^*_\theta\|\le r_0\}$, Step 1 gives $N\Phi_\theta(\Qh^*)=\frac1{2\beta}Z_N^\top W_\theta Z_N+R_N$ with
$|R_N|\le L_3\|Z_N\|^3/\sqrt N$, so $\E_\theta[|R_N|\mathbb 1_{A_N}]\le L_3L_1^3\E_\theta\|\Delta_N\|^3/\sqrt N\le c_2/\sqrt N$. On
$A_N^c$, $0\le N\Phi_\theta(\Qh^*)\le N(H+H^2)$ (as in the proof of Theorem~\ref{thm:asymp}(ii)); since
$\|\Qh^*-Q^*_\theta\|\le H\max_z|(\Ph-P_\theta)V^*_\theta(z)|$ and $V^*_\theta$ takes values in $[0,H]$, Hoeffding's
inequality gives $\Pr_\theta(A_N^c)\le2Ke^{-2Nr_0^2/H^4}$. With
$\E_\theta[Z_N^\top W_\theta Z_N\mathbb 1_{A_N^c}]\le\|W_\theta\|L_1^2(\E_\theta\|\Delta_N\|^4)^{1/2}\Pr_\theta(A_N^c)^{1/2}$, the event $A_N^c$
contributes at most $c_3/\sqrt N$. All constants are uniform over $\theta\in U$, which proves the display.

\emph{Step 4: consequences.} $\theta\mapsto\mathcal C(\theta)$ is continuous on $U$, so
$\sup_{\|\theta-\theta_0\|\le c/\sqrt N}|\mathcal C(\theta)-\mathcal C(\theta_0)|\to0$ for every $c$; with the display this gives the
limit, and Theorem~\ref{thm:asymp}(iii) gives the matching lower bound.

\emph{Fixed allocations.} Let $n_z=\lfloor w_zT\rfloor$ samples be drawn at $z$, with $T\ge2/\min_zw_z$ so that
$n_z\ge w_zT/2\ge1$. The same steps apply with $N$ replaced by $T$ and
$\Delta_T=(\sqrt T(\Ph(\cdot\mid z)-P_\theta(\cdot\mid z)))_z$. Its covariance is the multinomial covariance times
$T/n_z=1/w_z+O(1/T)$, which changes $\E[L_TL_T^\top]$ by $O(1/T)$; the moment bounds hold with constants
that depend on $\min_zw_z$, and the Hoeffding exponents are multiplied by $\min_zw_z/2$. This gives the
display with $\mathcal C$ replaced by $\frac1{2\beta}\sum_zc_z/w_z$. The matching lower bound among estimators that
use this allocation follows from the first inequality in \eqref{eq:vantrees} with the fixed design
$\bar w=w$, in the same way as Theorem~\ref{thm:asymp}(iii) follows for $w_z=1/K$.
\end{proof}

%% file: app_misc.tex
\section{Exact inference: the jackpot MDP}\label{app:misc}

The ``jackpot'' MDP has a decision state $x$ with two actions and single-action states
$u,G,w,z$. Action $1$ moves to $u$, which moves to $G$ with probability $q$ and to $z$
otherwise; $G$ is absorbing with reward $1$, $z$ is absorbing with reward $0$; action $2$
moves to $w$, absorbing with reward $1/4$; all other rewards are $0$ and $\pref(\cdot\mid x)$
is uniform. The exact-inference recursion
$Q^{\rm EI}(s,a)=r(s,a)+\gamma\beta\log\E_{s'}e^{V^{\rm EI}(s')/\beta}$,
$V^{\rm EI}(s)=\beta\log\sum_a\pref(a\mid s)e^{Q^{\rm EI}(s,a)/\beta}$ gives
$Q^{\rm EI}(x,1)=\gamma^2\beta\log(1-q+qe^{H/\beta})$ and $Q^{\rm EI}(x,2)=\gamma H/4$; the soft-optimal
(variational) values are $Q^*(x,1)=\gamma^2qH$ and $Q^*(x,2)=\gamma H/4$.

\begin{proposition}\label{prop:ei}
Let $\gamma\ge3/4$, $H/\beta\ge32$ and $q_1=e^{-H/(2\beta)}$. Any estimator $\widehat p$ of the exact-inference
probability $\pi^{\rm EI}(1\mid x)$ that uses $N\le e^{H/(2\beta)}/3$ samples per pair has
$\max_{q\in\{0,q_1\}}\E_q|\widehat p-\pi^{\rm EI}_q(1\mid x)|\ge0.3$, while the variational
probabilities satisfy $|\pi^*_{0}(1\mid x)-\pi^*_{q_1}(1\mid x)|\le\gamma^2Hq_1/(4\beta)$.
\end{proposition}
\begin{proof}
Under $q=0$ the log-odds of action $1$ are $-\gamma H/(4\beta)$. Under $q_1$,
$Q^{\rm EI}(x,1)\ge\gamma^2\beta\log(q_1e^{H/\beta})=\gamma^2H/2$, so the log-odds are at least
$\gamma H(2\gamma-1)/(4\beta)\ge\gamma H/(8\beta)$. Hence the two exact-inference probabilities differ by at
least $\sigma(\gamma H/(8\beta))-\sigma(-\gamma H/(4\beta))\ge1-2e^{-\gamma H/(8\beta)}\ge0.9$. Only the samples at $u$ are
informative; the total-variation distance between their laws is
$1-(1-q_1)^N\le Nq_1\le1/3$. Le Cam's two-point argument gives
$\max_q\E_q|\widehat p-\pi^{\rm EI}_q|\ge\frac12\cdot0.9\cdot(1-\frac13)=0.3$. For the variational
policies the log-odds differ by $\gamma^2q_1H/\beta$ and the logistic function is
$\frac14$-Lipschitz.
\end{proof}

It is a special case of the exponential lower bounds for entropic-risk objectives
\citep{fei2020risk,mortensen2025recursive}, small enough to show the difference between the
exact and the variational posterior directly.

\begin{figure}[h]
\centering
\includegraphics[width=\linewidth]{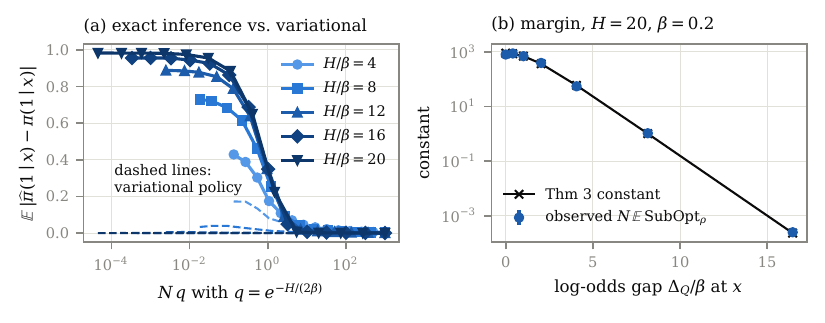}
\caption{(a) Jackpot MDP, $H=40$: mean absolute error of the plug-in estimate of
$\pi(1\mid x)$ against $Nq$ with $q=e^{-H/(2\beta)}$, for the exact-inference posterior (solid)
and the variational policy (dashed), 6000 datasets per point.
(b) Two-arm gadget, $H=20$, $\beta=0.2$, $N=10^6$: the constant of Theorem~\ref{thm:asymp}
(line) and $N\,\E\,\SubOpt$ (1200 datasets per point) against the log-odds gap at the
decision state. Standard errors are smaller than the markers.}
\label{fig:inference}
\end{figure}

\section{Simulation details}\label{app:experiments}

\CodeAvailability{} Soft-optimal policies are computed by soft policy iteration (exact evaluation
by a linear solve, then a softmax step) to a Bellman residual below $10^{-8}$ relative to $\|Q\|$. A
Monte Carlo estimate draws independent datasets with $N$ samples per pair and evaluates the plug-in
policy exactly in the true MDP. The captions give the number of datasets behind each estimate, and
error bars and $\pm$ values are one standard error.

Worst cases over the gap $\Delta$ are chosen on one batch of datasets and re-estimated on a fresh one
(three times, with 300 datasets per batch), to avoid the upward bias of maximizing noisy estimates.
When the first batch picks a $\Delta$ with lower risk, the re-estimate can come out below the value at
$\Delta=0$, as it does for some $H$ in Table~\ref{tab:horizon}. At $\Delta=0$ only the two loop pairs of the
gadget are random: with stay counts $k_a$, the plug-in policy at the decision state is the softmax of
$\gamma/(1-\gamma k_a/N)$ over $\beta$, and by Lemma~\ref{lem:pathkl} the risk is $\beta$ times its KL divergence
from the uniform policy. Table~\ref{tab:horizon} uses this closed form.

On the data of Figure~\ref{fig:rates}, the fitted slopes in $N$ average $-0.53$ in the slow
regime ($N\le0.03H^3/\beta^2$, $\beta\in\{0.003,0.03,0.3\}$) and are $-1.01$ in the fast one
($N\ge30H^3/\beta^2$, $\beta=3$). Starting the fast-regime window at $3$ or $10$ times $H^3/\beta^2$ gives slopes
no flatter than $-0.98$. For the time-indexed gadget we compute the constant of
Theorem~\ref{thm:asymp} in closed form for $20\le H\le160$; a Monte Carlo estimate at $H=20$ is
$1.01$ times the closed form. In Figure~\ref{fig:inference}(a), at $N\approx e^{H/(2\beta)}$ the error of the
exact-inference estimate is still at least $0.23$ times the gap between the two
hypotheses. All simulations together take about 18 minutes on a 4-core CPU.

\paragraph{Instances.} The two-arm gadget is the MDP of Theorem~\ref{thm:lower}(a) with $p_0=(4\gamma-1)/(3\gamma)$
and $p_{1,2}=p_0\pm\Delta$. Its time-indexed version replaces each loop by a chain of $\lceil8H\rceil$ distinct
states with reward $1$ that move to the next state of the chain with probability $p_0$ and otherwise to the
terminal state. Garnet instances have $S$ states, $A$ actions and branching $b$: each pair moves to $b$
uniformly chosen states with Dirichlet$(1)$ weights, rewards are uniform on $[0,1]$, the rows of $\pref$ are
Dirichlet$(1)$ and $\rho$ is uniform. The ring has 100 states on a cycle and two actions; action $a$ moves
$a+1$ steps forward with probability $0.7$ and $a+2$ steps with probability $0.3$, rewards are uniform on
$[0,1]$, $\pref$ is uniform and $\rho$ is a point mass. The allocation experiment uses the gadget ($H=20$,
$\beta=1$, $\Delta=0$) and Garnets with $(S,A)=(20,2)$ and $(10,4)$, $b=3$, $\gamma=0.95$ and $\beta=0.5$; pairs with
$c_z=0$ get one sample.

\begin{table}[h]
\caption{Horizon dependence. Fast regime: $\beta N\,\E\,\SubOpt_\rho$ with $\beta=1$ and $N=256H^3$, at the
worst gap $\Delta$ (900 datasets) and at $\Delta=0$ ($3\cdot10^6$ datasets), next to the constant of
Theorem~\ref{thm:asymp}. Slow regime: worst case over $\Delta$ of $\E\,\SubOpt_\rho/\sqrt{H^3/N}$ with
$\beta=10^{-4}$ and $N=64H$ (900 datasets). The $\pm$ values are one standard error.}
\label{tab:horizon}
\centering\small
\begin{tabular}{rrrrr}
\toprule
$H$ & fast, worst case over $\Delta$ & fast, $\Delta=0$ & Thm.~\ref{thm:asymp} constant & slow \\
\midrule
5 & 1.424 $\pm$ 0.066 & 1.5458 $\pm$ 0.0013 & 1.5469 & 0.047 $\pm$ 0.003 \\
10 & 16.31 $\pm$ 0.78 & 18.505 $\pm$ 0.015 & 18.51 & 0.054 $\pm$ 0.004 \\
20 & 184.5 $\pm$ 9.5 & 177.79 $\pm$ 0.15 & 177.68 & 0.059 $\pm$ 0.004 \\
40 & 1561 $\pm$ 77 & 1551.1 $\pm$ 1.3 & 1550.7 & 0.071 $\pm$ 0.005 \\
80 & 13200 $\pm$ 600 & 12940 $\pm$ 11 & 12945 & 0.071 $\pm$ 0.004 \\
160 & 103700 $\pm$ 4800 & 105720 $\pm$ 86 & 105770 & 0.071 $\pm$ 0.004 \\
320 & 808500 $\pm$ 39000 & 855260 $\pm$ 700 & 855030 & 0.075 $\pm$ 0.004 \\
\bottomrule
\end{tabular}
\end{table}

\begin{table}[h]
\caption{The constant $\frac1{2\beta}\mathrm{tr}(W\Sigma_Q)$ of Theorem~\ref{thm:asymp} and the measured
$N\,\E\,\SubOpt_\rho$ (3000 datasets, $\pm$ one standard error).}
\label{tab:constants}
\centering\footnotesize
\begin{tabular}{lrrrr}
\toprule
instance & $N$ & constant & measured & ratio \\
\midrule
gadget, $H=20$, $\beta=1$, $\Delta=0$ & $10^4$ & 177.7 & 168.1 $\pm$ 4 & 0.946 \\
gadget, $H=20$, $\beta=1$, $\Delta=0$ & $10^5$ & 177.7 & 177.9 $\pm$ 4.6 & 1.001 \\
gadget, $H=20$, $\beta=1$, $\Delta=0$ & $10^6$ & 177.7 & 177.3 $\pm$ 4.6 & 0.998 \\
gadget, $H=20$, $\beta=1$, $\Delta=0.002$ & $10^4$ & 150.5 & 145.1 $\pm$ 3.5 & 0.964 \\
gadget, $H=20$, $\beta=1$, $\Delta=0.002$ & $10^5$ & 150.5 & 148.7 $\pm$ 3.8 & 0.988 \\
gadget, $H=20$, $\beta=1$, $\Delta=0.002$ & $10^6$ & 150.5 & 153.7 $\pm$ 4.1 & 1.021 \\
gadget, $H=40$, $\beta=3$, $\Delta=0$ & $10^4$ & 516.9 & 487.7 $\pm$ 12 & 0.944 \\
gadget, $H=40$, $\beta=3$, $\Delta=0$ & $10^5$ & 516.9 & 513.5 $\pm$ 14 & 0.993 \\
gadget, $H=40$, $\beta=3$, $\Delta=0$ & $10^6$ & 516.9 & 520.3 $\pm$ 13 & 1.007 \\
Garnet, $S=10$, $A=3$, $\gamma=0.9$, $\beta=0.1$ & $10^4$ & 0.5167 & 0.5209 $\pm$ 0.0053 & 1.008 \\
Garnet, $S=10$, $A=3$, $\gamma=0.9$, $\beta=0.1$ & $10^5$ & 0.5167 & 0.5103 $\pm$ 0.0051 & 0.988 \\
Garnet, $S=10$, $A=3$, $\gamma=0.9$, $\beta=0.1$ & $10^6$ & 0.5167 & 0.5153 $\pm$ 0.0052 & 0.997 \\
Garnet, $S=10$, $A=3$, $\gamma=0.9$, $\beta=1$ & $10^4$ & 0.04478 & 0.04481 $\pm$ 0.00034 & 1.001 \\
Garnet, $S=10$, $A=3$, $\gamma=0.9$, $\beta=1$ & $10^5$ & 0.04478 & 0.04415 $\pm$ 0.00034 & 0.986 \\
Garnet, $S=10$, $A=3$, $\gamma=0.9$, $\beta=1$ & $10^6$ & 0.04478 & 0.0445 $\pm$ 0.00033 & 0.994 \\
Garnet, $S=30$, $A=4$, $\gamma=0.95$, $\beta=0.5$ & $10^4$ & 0.1442 & 0.1443 $\pm$ 0.00057 & 1.001 \\
Garnet, $S=30$, $A=4$, $\gamma=0.95$, $\beta=0.5$ & $10^5$ & 0.1442 & 0.1435 $\pm$ 0.00057 & 0.995 \\
Garnet, $S=30$, $A=4$, $\gamma=0.95$, $\beta=0.5$ & $10^6$ & 0.1442 & 0.1449 $\pm$ 0.00056 & 1.005 \\
ring, $S=100$, $\gamma=0.95$, $\beta=0.5$ & $10^4$ & 0.1441 & 0.1439 $\pm$ 0.0007 & 0.999 \\
ring, $S=100$, $\gamma=0.95$, $\beta=0.5$ & $10^5$ & 0.1441 & 0.1441 $\pm$ 0.00071 & 0.999 \\
ring, $S=100$, $\gamma=0.95$, $\beta=0.5$ & $10^6$ & 0.1441 & 0.1451 $\pm$ 0.00071 & 1.006 \\
\bottomrule
\end{tabular}
\end{table}